\documentclass[11pt]{article}

\usepackage[preprint]{acl}

\usepackage{times}
\usepackage{latexsym}
\usepackage{wrapfig}
\usepackage[T1]{fontenc}
\usepackage[utf8]{inputenc}
\usepackage{microtype}
\usepackage{inconsolata}
\usepackage{graphicx}
\usepackage{microtype}
\usepackage{amssymb}
\usepackage{amsmath}
\usepackage{hyperref}
\usepackage{url}
\usepackage{booktabs}
\usepackage{xcolor}
\usepackage{dsfont}
\usepackage{subcaption}
\usepackage{float}
\usepackage{algorithm}
\usepackage{algorithmic}
\usepackage[most]{tcolorbox}
\usepackage{xcolor}
\usepackage{epigraph}
\usepackage{graphicx}
\usepackage{tcolorbox}
\usepackage{color, soul}
\usepackage{wrapfig}
\definecolor{gboxcolor}{gray}{0.9} 
\sethlcolor{gboxcolor}

\newtcbox{\gbox}{
  on line,
  colback=gray!20, 
  colframe=gray!20, 
  boxrule=0pt,      
  arc=3pt,          
  boxsep=0pt,
  left=2pt, right=2pt, top=1.5pt, bottom=1.5pt
}

\definecolor{takeawaybg}{RGB}{168, 197, 218}
\definecolor{takeawayborder}{RGB}{45, 122, 177}

\newtcolorbox{takeawaybox}[1]{
  colback=takeawaybg,
  colframe=takeawayborder,
  boxrule=0.7pt,
  arc=2pt,
  left=6pt,
  right=6pt,
  top=6pt,
  bottom=6pt,
  title=\textit{#1},
  fonttitle=\bfseries,
  before skip=0.5cm,
  after skip=0.5cm
}

\newcommand\blfootnote[1]{%
  \begingroup
  \renewcommand\thefootnote{}\footnote{#1}%
  \addtocounter{footnote}{-1}%
  \endgroup
}

\usepackage{lineno}

\definecolor{darkblue}{rgb}{0, 0, 0.5}
\hypersetup{colorlinks=true, citecolor=darkblue, linkcolor=darkblue, urlcolor=darkblue}

\title{Demystifying Reinforcement Learning Post-Training of \\Language Models}

\author{Donovan Clay$^{1,*}$,
Saket Gollapudi$^{1,*}$,
Sankar Harilal$^{1,*}$,
Min Jang$^{1,*}$,
Jacob Morrison$^{1,2}$, \\
\textbf{Sewoong Oh}$^{1}$, 
\textbf{Natasha Jaques$^{1}$} \\
\small $^{*}$ Equal contributions \\
\small $^{1}$ University of Washington \\
\small $^{2}$ Allen Institute for AI
}

\begin{document}
\maketitle
\begin{abstract}
Reinforcement learning (RL) post-training has emerged as a powerful framework for enhancing the capabilities of large language models (LLMs), enabling impressive reasoning, math, and coding capabilities. Yet for many researchers and practitioners, the principles behind classical RL remain a ``black box''. In this work, we deconstruct the RL post-training algorithm, investigating each step to clarify what is actually happening beneath the surface. By isolating the mechanics of RL with Verifiable Rewards in a controlled and simplified environment, we examine how RL outcomes are shaped by the base model’s prior distribution, the granularity of the reward signal, the diversity of the prompt distribution, and model scale. We use the entropy of the policy's output distribution as a lens to compare the distributions learned through pretraining, SFT, and RL post-training, revealing how each stage shapes model certainty. Our investigation sheds light on how these choices interact to affect post-training success. For example, we show that the effect of so-called `spurious rewards' depends on the prompt distribution used for post-training. We also provide insight into why the success of RL post-training depends on whether the base model already places sufficient probability mass on the desired behavior, linking it to the classical concept of exploration in RL. Ultimately, we provide this primer as a resource to those in the NLP community wishing to incorporate RL as a tool in their toolbox.
\blfootnote{Website: \href{https://minjang10.github.io/demystifying-rl-finetuning-web/}{ https://minjang10.github.io/demystifying-rl-finetuning-web/} \\
Code: \href{https://github.com/sankarh-1/demystifying-rl-finetuning}{https://github.com/sankarh-1/demystifying-rl-finetuning}}
\end{abstract}

\section{Introduction}

Reinforcement Learning (RL) has rapidly ascended as a dominant framework for post-training Large Language Models (LLMs), proving essential for aligning models with human intent, enhancing reasoning capabilities, and ensuring safety \citep{ouyang2022traininglanguagemodelsfollow, Guo_2025, hu2024openrlhf}.  

However, its rapid rise in popularity, combined with the fact that understanding RL post-training for LLMs requires deep expertise in both classical RL and NLP, has led to several misunderstandings about its mechanisms and capabilities in literature.

\begin{figure*}[t]
   \centering
   \includegraphics[width=\linewidth]{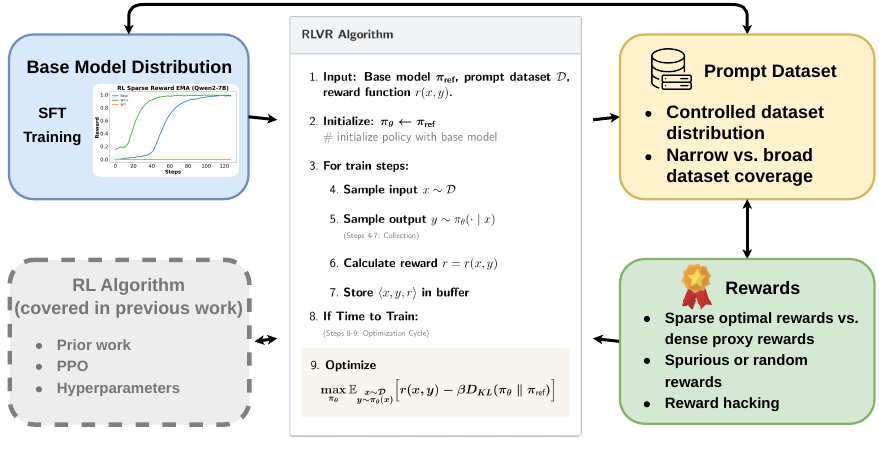}
   \caption{Overview of the RL Post-Training Process. Our experiments isolate the specific impact of the Base Distribution, the Prompt Distribution ($\mathcal{D}$), and the Reward Function ($r$). We exclude algorithmic variation from our scope, deferring to the extensive existing research on RL algorithmic choices and engineering techniques that enhance performance (see e.g. \citep{khatri2025artscalingreinforcementlearning})}
   \label{fig:RL-overview}
\end{figure*}

We offer this paper as a deep dive into the mechanics of RL post-training. Figure~\ref{fig:RL-overview} gives an overview of the post-training algorithm. At a high level, it is an iterative process that shapes an LLM's output distribution using reward signals that might not be differentiable. Given a prompt, responses are sampled from the model, scalar rewards are assigned based on their quality, and model parameters are adjusted to encourage higher-reward responses. Our goal is to study each component of this algorithm through carefully controlled experiments, showing how manipulating the base model distribution, prompt distribution, and reward function all induce changes in the success rate of post-training a model in order to achieve a particular goal. 

Some of the conclusions we draw may seem obvious to researchers already well-versed in RL. However, we point out that several aspects of RL post-training of LLMs make it distinct from classical RL, and require special attention. 
First, RL post-training does not need to explore from scratch to discover highly rewarding behaviors, and instead leverages strong prior knowledge. The policy being learned is initialized with a base language model and is constrained to stay close to this base reference policy, in order to prevent language drift \citep{jaques2017sequencetutorconservativefinetuning}. As we will show, the base model's distribution exerts strong effects on post-training outcomes. Second, post-training attempts to modify token-level probabilities, typically using utterance or turn-level rewards. Optimizing behavior over a multi-turn conversation could easily involve sequence $10,000+$ actions, whereas past high-profile successes in classical RL were limited to 250-3000 actions \cite{silver2016mastering,vinyals2019grandmaster}. Thus post-training represents an application of RL at an unprecedented scale, sometimes exhibiting surprising and unintuitive results due to the complexities of the underlying language models \cite{jorgenvaag2026reinforcement}.

Because of the KL-constraint, post-training of LLMs can be understood as redistributing probability mass within the pre-trained distribution. To analyze these distributional shifts, we track the entropy of the policy throughout training, as well as the probability mass assigned to certain behaviors of interest, which allows us to directly compare how pretraining, SFT, and DPO each shape the model's output distribution and how resilient each is to corruption. 

While prior work has studied \textit{what} RL learns through the redistribution of probability mass to base model behaviors \cite{zhao2025echo,rajani2025scalpel,yang2026beyond,akgul2026rethinking}, our goal is to study \textit{why} and \textit{when} this happens, by experimentally manipulating the base model distribution. 

To demonstrate these phenomena, we use a controlled sequence-generation setting where probability shifts and task success can be measured exactly, utilizing toy RLVR tasks (producing a target string, and solving a math problem) to systematically vary the base model's pre-training distribution. Using SFT, we increase the probability of the target string in the base model (SFT+), and also negate the process to disrupt this knowledge from the base model (SFT-). Across models of different scales, we analyze how these interventions affect post-training under different reward signals, including sparse, dense, and random rewards, as well as varying prompt distributions, including on narrow reasoning datasets, and the broad pre-training distribution used for Tulu-3 \citep{lambert2025tulu3pushingfrontiers}. Although simplified, this setup enables precise analysis of how factors like the base model's distribution affect outcomes, an analysis that would be impossible in full-scale LLM post-training. Taken together, these results provide a clearer picture of the mechanisms behind RL post-training. We hope this paper will serve as a useful primer for those wishing to gain more expertise in this domain, or deepen their understanding of how to develop more effective post-training pipelines. 

\section{Related Work}

RL post-training has proven highly effective for LLM reasoning and alignment \citep{Guo_2025, lambert2025tulu3pushingfrontiers, ouyang2022traininglanguagemodelsfollow}. However, recent works have surfaced seemingly contradictory findings and misconceptions that our experiments aim to clarify.

For example, \citet{shao2025spuriousrewardsrethinkingtraining} showed a surprising result: RL with spurious (random or incorrect) rewards can still produce performance gains on math benchmarks for a certain base model family. This has led some to suspect that many post-training successes could be attributable to this `spurious' phenomenon. In contrast, our framework shows that this effect is governed by the prompt distribution: restricting training to a narrow prompt distribution can improve performance within that domain (if the model already assigns high probability to the correct answer), whereas training with random rewards over a broad prompt distribution increases policy entropy and degrades capabilities.

Meanwhile, questions have arisen as to whether RL is capable of training the model to learn new behaviors, or whether it merely upweights correct behaviors already present in the base model \citep{yue2025doesreinforcementlearningreally}. Excellent work has shown that offline methods like DPO, which operate on a fixed dataset, tend to concentrate probability mass on responses that already had high likelihood in the base model \citep{ren2025learningdynamicsllmfinetuning}. But is this true of online RL that explores to generate new data as it trains?

Prior work has studied what RL learns through the redistribution of probability mass to base model behaviors \cite{zhao2025echo,rajani2025scalpel,yang2026beyond,akgul2026rethinking}. These findings describe the outcome but not its cause; we instead manipulate the base model distribution directly to isolate why and when this redistribution occurs. We find, unsurprisingly, that the probability of sampling a given response in the base model strongly affects how easily that response can be learned with post-training, since the base model's distribution affects what responses are sampled during post-training, and thus the quality of exploration needed for RL. In other words, behaviors with negligible initial probability are difficult to discover and reinforce, a perspective that aligns with theoretical frameworks such as the ``Coverage Principle'' 
\citep{chen2025coverageprinciplepretrainingenables}. 

However, we show that this limitation is strictly bound to settings with sparse rewards. \textit{With a sufficiently dense and correct reward function, RL post-training can succeed in teaching a model new behaviors that have negligible support in the base model}. While this result is well-known in classical RL \cite{eysenbach2018diversity}, it has been heavily called into question in recent post-training literature \cite{yue2025doesreinforcementlearningreally,shao2025spuriousrewardsrethinkingtraining,wu2507invisible,zhao2025echo,wu2025position}. By systematically varying these components, our work reconciles these seemingly contradictory findings. For a comprehensive literature review, see Appendix \ref{app:expanded_related_work}.

\section{Background: RL for Language Models}

We formulate the text generation process as a finite horizon Markov decision process (MDP), defined as the tuple $\mathcal{M} = \langle \mathcal{S}, \mathcal{A}, \mathcal{T}, r, T \rangle$. Here, $\mathcal{S}$ is the set of possible states, $\mathcal{A}$ is the set of possible actions, $r: \mathcal{S} \mapsto \mathbb{R}$ is the reward function, and $T \in \mathbb{Z}_{+}$ is the finite planning horizon depth. The transition function is $\mathcal{T}: \mathcal{S} \times \mathcal{A} \times \mathcal{S} \mapsto [0, 1]$, which defines the probability of reaching state $s_{t+1}$ given action $a_t$ and state $s_t$. Note that for an LLM deployed in an interactive dialog with a human user, $\mathcal{T}$ depends on the human's policy and response. Therefore, to enable precise analysis, we restrict our focus to the RLVR setting, where we are attempting to learn an auto-regressive LLM policy ($\pi_\theta$), and we map these definitions as follows:

\begin{itemize}
\itemsep-0.1mm
    \item \textbf{State ($\mathcal{S}$)}: All possible sequences of tokens up to length $T$ (the prompt $x$ plus the tokens of $y$ generated so far).
    \item \textbf{Action ($\mathcal{A}$)}: The model's vocabulary (every possible next token).
    \item \textbf{Transition ($\mathcal{T}$)}: A deterministic step that appends the next token to the current context, i.e., $s_{t+1} \leftarrow s_t \oplus a_t$ (where $\oplus$ is the concatenation operator).
    \item \textbf{Horizon ($T$)}: The model's maximum token generation limit.
    \item \textbf{Reward function $r(s)$}: A deterministic verifier applied to the generated sequence (e.g., binary correctness for math). 
    
\end{itemize}
While this formulation aligns with the classical RL view of sequential decision-making over $\mathcal{M}$, the structure of text generation induces a significantly simplified, deterministic system, with verifiable rewards assigned at the sequence level, rather than at intermediate steps. From an RL perspective, this corresponds to the simplified setting of a contextual bandit; 
through an NLP lens, the problem is typically viewed as learning a probability distribution over complete sequences, where RL post-training effectively reshapes this distribution in response to reward signals. 

\textbf{Process Reward Models (PRMs):} 
evaluate intermediate reasoning steps and award partial credit for logical milestones, enabling 
step-level credit assignment. As we will show, this can enable iteratively reshaping the model's output distribution to explore behaviors that have low probability in the base model, and hence would not be explored or learned in a sparse reward setting.

\textbf{Policy learning:}
During post-training, a language model acts as a stochastic policy $\pi_\theta(y|x)$, parameterized by weights $\theta$. The model generates a sequence of tokens $y = (y_1, y_2, ..., y_T)$ given a prompt $x$ drawn from a dataset $\mathcal{D}$. The objective is to maximize the expected reward $r(x, y)$ while maintaining proximity to a reference policy $\pi_{\text{ref}}$ (typically the initial SFT model) to prevent reward hacking and degradation of language fluency \citep{jaques2017sequencetutorconservativefinetuning}.  
Formally, the objective is:
\begin{equation*}
    U(\pi_\theta) = \mathbb{E}_{x\sim {\cal D},y\sim \pi_\theta} \Big[\frac{1}{c}r(x,y) - D_{\rm KL}( \pi_\theta \| \pi_{\rm ref})\Big] 
\end{equation*}
where ${\cal D}$ is a dataset of prompts, $D_{KL}$ denotes the KL divergence, $c$ is a scaling constant, and $\pi_\theta=\pi_\theta(\cdot|x)$ and $\pi_{\rm ref} = \pi_{\rm ref}(\cdot|x)$ are conditioned on the prompt $x$.

\section{The Mechanics of RL Post-Training}

The RL post-training algorithm (Figure \ref{fig:RL-overview}) relies on three components: a base model, a reward function, and a curated prompt dataset. To systematically investigate how each element impacts learning dynamics, we conduct two targeted experimental phases in a controlled NLP sandbox. The first phase isolates the interaction between the base model distribution and the reward function by training the model to reliably generate a specific target string $\tau$ (Sections \ref{sec:base_distribution}-\ref{sec:reward_shaping}). Phase 2 (Section \ref{sec:prompt_distributions}) examines how the effect of training on spurious rewards is modulated by the prompt dataset. 

\subsection{Experimental Design: Manipulating the Base Distribution}
\label{sec:base_distribution}
To isolate how a model's learned base distribution shapes RL dynamics 
(Figure \ref{fig:RL-overview}, Line 2),
we systematically manipulate the initial probability of a target behavior in the base model. We evaluate this across two distinct settings: a controlled sequence-generation sandbox and a combinatorial mathematical reasoning environment. Across both tasks, we establish three distinct initialization variants to precisely measure how the reference policy's initial coverage dictates RL success. The \textbf{Base} model is the unmodified base model (we test OLMo 3, Qwen2-7B, Qwen3-1.7B, Qwen2.5-7B-Instruct). 

The \textbf{SFT+} model is the base model fine-tuned on the correct target behavior to artificially inflate the prior probability of the desired sequence. The \textbf{SFT-} model is the base model fine-tuned on the opposite objective: to 

\textit{maximize} cross-entropy loss on the target behavior to suppress the probability of sampling the target behavior to nearly 0\%.

\textbf{Target string generation.} In our sequence-generation sandbox, we focus on the target behavior of producing an exact string $\tau$, making it possible to precisely measure the probability of this behavior in a particular model, and trivially verify RL success. We start with producing the movie quote 
$\tau=$ \textit{``life is like a box of chocolates you never know what youre gonna get''}.
To obtain base models with different probabilities of producing $\tau$, 
we use SFT+/- where the model is trained to produce the target quote $\tau$ 20\% of the time, and on other random quotes sampled from the Cornell Movie-Dialogs Corpus \cite{Danescu-Niculescu-Mizil2011} on 80\% of samples, to prevent catastrophic forgetting. 

\textbf{RLVR reasoning.} To demonstrate that the principles of coverage extend to complex problem solving, we scale this setup to multi-step mathematical reasoning using a 
problem from the AIME 2025 dataset (Problem 4): \textit{Find the number of ordered pairs $(x,y)$, where both $x$ and $y$ are integers between $-100$ and $100$ inclusive, such that:$12x^2 - xy - 6y^2 = 0$}. We can then assess the base model's probability of achieving the right answer (117) across several samples (128), and once again experimentally manipulate this probability using similar SFT+/- training on the correct answer.

\subsection{Reward Function: from Sparse to Dense}
\label{sec:reward_shaping}
A key challenge in applying RLVR is the sparsity of reward signal used to update the policy (Figure \ref{fig:RL-overview}, Line 6). In verifiable tasks, rewards are typically binary, and applied only at the sequence level. For our sequence-generation task, $r_{\text{sparse}}(y) = \mathds{1}\{ \tau \subseteq y\}$, with a length penalty applied to prevent reward hacking via ``rambling'' \citep{chevalierboisvert2023minigridminiworldmodular}. For the math reasoning task, the sparse reward remains a strict binary verifier, assigning 1 if the final extracted answer exactly matches the ground truth, and 0 otherwise. 

Strict sparsity creates a severe exploration bottleneck: if the reference policy $\pi_{\theta}$ assigns near-zero probability to the target, the model may never produce $\tau$, and thus receive a learning signal. This mirrors the exploration challenge in systems like DeepSeek-R1 \citep{Guo_2025}, where models receive zero reward unless a long chain-of-thought yields the exact correct answer. 

\textbf{Dense rewards: Edit Distance.} Classical RL frequently trains randomly initialized policies to perform complex, novel tasks \citep{christiano2023deepreinforcementlearninghuman, xie2025kungfubotphysicsbasedhumanoidwholebody, cheng2023extremeparkourleggedrobots}. Thus, we hypothesize that LLMs can similarly learn behaviors with vanishingly small base probabilities, if provided with dense rewards (reward shaping), which can mitigate exploration bottlenecks by providing feedback on incremental approximations. Thus, we implement dense reward proxies for both tasks. For sequence generation, we use Levenshtein distance as a proxy for dense feedback:  
\vspace{-6pt}
\begin{equation}
\small
D(i,j)=
\begin{cases}
\max(i,j), & \text{if } i \text{ or } j=0,\\[4pt]
\min\left\{
\begin{aligned}
&D(i-1,j)+1,\\
&D(i,j-1)+1,\\
&D(i-1,j-1)\\
&\qquad+\mathbb{1}(y_i\neq\tau_j)
\end{aligned}
\right\}, & \text{otherwise.}
\end{cases}
\end{equation}

While it does not provide a per-token reward along the trajectory, it offers intermediate gradient signals based on structural proximity to the target (Refer to Appendix \ref{app:reward_implementations} for additional detail).

\textbf{Dense rewards: Process Reward Model.} For mathematical reasoning, we replace structural distance with a Process Reward Model (PRM) with further details in Appendix \ref{app:prm_details}. Utilizing an LLM-as-a-judge, the PRM evaluates intermediate reasoning steps, awarding partial credit for discovering any of five key valid sub-components of the ground-truth solution. 
While PRMs have shown mixed results in prior literature \citep{lightman2023letsverifystepstep, setlur2024rewardingprogressscalingautomated, tiwari2026rewardattackanalyzingrobustness, Guo_2025}, our idealized setup examining only a single problem enables us to construct an accurate, rich, and dense reward. We can then test whether a model can be guided through a combinatorial space to construct a novel reasoning chain that has near-zero support in its original base model's distribution.

\subsection{The Role of Prompt Distributions}
\label{sec:prompt_distributions}
Recent work shows that RL post-training with random rewards can improve reasoning capabilities, but only for the Qwen family of models \citep{shao2025spuriousrewardsrethinkingtraining}. We hypothesize that this effect is fundamentally governed by not only the base model's distribution, but also the prompt distribution $\mathcal{D}$, which determines which states the model explores during RL training (Figure~\ref{fig:RL-overview}). 
For any prompt, highly probable responses under the base model are sampled more frequently and thus disproportionately rewarded by chance.
For the special case of training on a narrow distribution of prompts $\mathcal{D}$ (e.g., exclusively math), where the base model already has a high probability of sampling the correct answer, even an uninformative, random reward can further reinforce those answers, decreasing policy entropy and potentially increasing capabilities.
Conversely, training on a broad $\mathcal{D}$ uniformly rewards arbitrary behaviors, increasing overall policy entropy and inducing global unlearning. In other words, if the reward equally incentivizes all behaviors, the policy may eventually converge to a near-uniform token distribution, which would not comprise a capable model with good performance.

To test these hypotheses, we train both Qwen2.5-Math-7B and OLMo 3 base models with random rewards drawn uniformly from $[0, 1]$.  
Two base models with contrasting priors let us separate the effect of $\mathcal{D}$ from the base model's existing bias. Qwen2.5-Math-7B carries a strong math prior and reproduces the spurious-rewards phenomenon; OLMo 3 carries no such prior and its evaluation scores do not artificially improve under random rewards \citep{olmo3-report}. For Qwen, the broad distribution is drawn from the WildChat dataset \citep{zhao2024wildchat1mchatgptinteraction} (10k prompts) and the narrow distribution is drawn from DeepScaleR \citep{tan2026deepscaler} (100 prompts). For OLMo, the broad distribution condition consists of 10k mixed prompts from the OLMo~3 RLVR training mix and the narrow distribution condition consists of 100 math-only prompts. We also examine how post-training starting from three OLMo 3 base models (Base, SFT, and DPO) affects results using identical random reward training.
Using three successive checkpoints allows us to test a secondary hypothesis: progressive fine-tuning reduces the model's initial output entropy, making it more resilient to the entropy-increasing effects of random rewards. To evaluate, we track the log-likelihood and average per-token entropy of both in-distribution and held-out prompts throughout training, alongside a suite of reasoning benchmarks. Complete hardware, algorithmic, and evaluation sampling details are deferred to Appendix \ref{app:prompt_dist_setup}.

\section{Results: Unpacking Learning Dynamics}

\begin{table}[b]
    \centering
    \begin{subtable}{0.48\textwidth}
        \centering
        \footnotesize
        \begin{tabular}{lrrr}
            \toprule
            \textbf{Model} & \textbf{SFT+} & \textbf{Base} & \textbf{SFT-} \\
            \midrule
            Qwen3-8B & 42.8\% & 0.00\% & 0.00\% \\
            Qwen2-7B & 28.75\% & 3.52\% & 0.00\% \\
            Qwen3-1.7B & 14.04\% & 0.48\% & 0.00\% \\
            Qwen2-1.5B & 7.96\% & 0.31\% & 0.00\% \\
            \bottomrule
        \end{tabular}
        \caption{Pre-RL probabilities.}
        \label{tab:pre-training-dist-prior}
    \end{subtable}
    \hfill
    \begin{subtable}{0.48\textwidth}
        \centering
        \footnotesize
        \begin{tabular}{lrrr}
            \toprule
            \textbf{Model} & \textbf{SFT+} & \textbf{Base} & \textbf{SFT-} \\
            \midrule
            Qwen3-8B & 100.0\% & 0.00\% & 0.00\% \\
            Qwen2-7B & 99.6\% & 99.9\% & 0.00\% \\
            Qwen3-1.7B & 98.7\% & 10.0\% & 0.00\% \\
            Qwen2-1.5B & 92.5\% & 0.71\% & 0.00\% \\
            \bottomrule
        \end{tabular}
        \caption{Post-RL probabilities (sparse rewards).}
        \label{tab:pre-training-dist-post}
    \end{subtable}
    \caption{Probability of producing the target quote before and after RL post-training, (eval = 10,000 samples).}
    \label{tab:pre-training-dist}
\end{table}

\subsection{The Coverage Principle}
The coverage principle proposes that pre-training enables post-training primarily by assigning non-negligible probability mass to high-quality data \citep{chen2025coverageprinciplepretrainingenables}. To test this empirically, we manipulated base model priors through SFT+ and SFT- phases, resulting in models that exhibit differing probabilities of producing the target string $\tau$, which are shown in Table \ref{tab:pre-training-dist-prior}. 

When training these models utilizing a standard sparse reward, we indeed find that the base model's probability of producing the desired behavior has a strong effect on the success of RL post-training.
Models artificially injected with the target quote (SFT+) demonstrated quick convergence and easily maximized the sparse reward. Conversely, models lacking coverage completely failed to learn (SFT- and the unmodified Qwen3-8B Base). Because the target quote was absent from the reference policy, it was never generated during the RL phase, trapping the agent in a state of zero reward. This failure provides empirical evidence to support the coverage principle: without initial probability mass, RL with sparse rewards does not succeed in learning a new behavior outside of the base model's original distribution (see Figure \ref{fig:qwen2-7b_Apdx} in the Appendix for Qwen2-7B learning curves). 

Furthermore, unmodified base models highlight the strict threshold of sparse RL. While Qwen2-7B ($\approx 3.5\%$ prior) eventually converged after a 40-step exploration plateau, smaller models with lower priors, like Qwen3-1.7B, ($\approx 0.5\%$ prior) failed entirely. With rewards so sparse, hitting the correct sequence by chance was statistically improbable under sparse optimization (Figure \ref{fig:dense_overlay_1b}).

We observe the same exploration bottleneck in our mathematical reasoning task (Table \ref{tab:pre-training-dist-aime}). Evaluating Qwen2.5-7B-Instruct on AIME Problem 4 with a sparse reward reveals similar behavior. Models injected with the correct derivation (SFT+) quickly converged ($\sim$85\% exact match), while the unmodified Base model stalled near 10\%. Because the complete reasoning chain was unlikely under the base distribution, it was rarely sampled during exploration, leaving the model with almost no positive learning signal. This confirms that the Coverage Principle governs real-world RLVR: without sufficient initial probability mass on complete reasoning trajectories, sparse optimization stalls.

\begin{table}[ht]
    \centering
    \begin{subtable}{0.48\textwidth}
        \centering
        \footnotesize
        \begin{tabular}{lccc}
            \toprule
            \textbf{Configuration} & \textbf{SFT+} & \textbf{Base} & \textbf{SFT-} \\
            \midrule
            No Reward & 26.6\% & 3.92\% & 0.00\% \\
            Sparse Reward & 85.9\% & 10.2\% & 0.00\% \\
            Dense Reward & 86.7\% & 92.2\% & 0.00\% \\
            \bottomrule
        \end{tabular}
        \label{tab:pre-training-dist-post-aime}
    \end{subtable}
    \caption{Probability of correctly solving AIME Problem 4 before and after RL post-training on Qwen2.5-7B-Instruct (eval = 128 samples).}
    \vspace{-12pt}
    \label{tab:pre-training-dist-aime}
\end{table}

\subsection{Dense Reward Shaping: Beyond Pass@k}

\begin{figure*}[ht]
    \centering
        \begin{subfigure}{0.49\linewidth}
        \includegraphics[width=\linewidth]{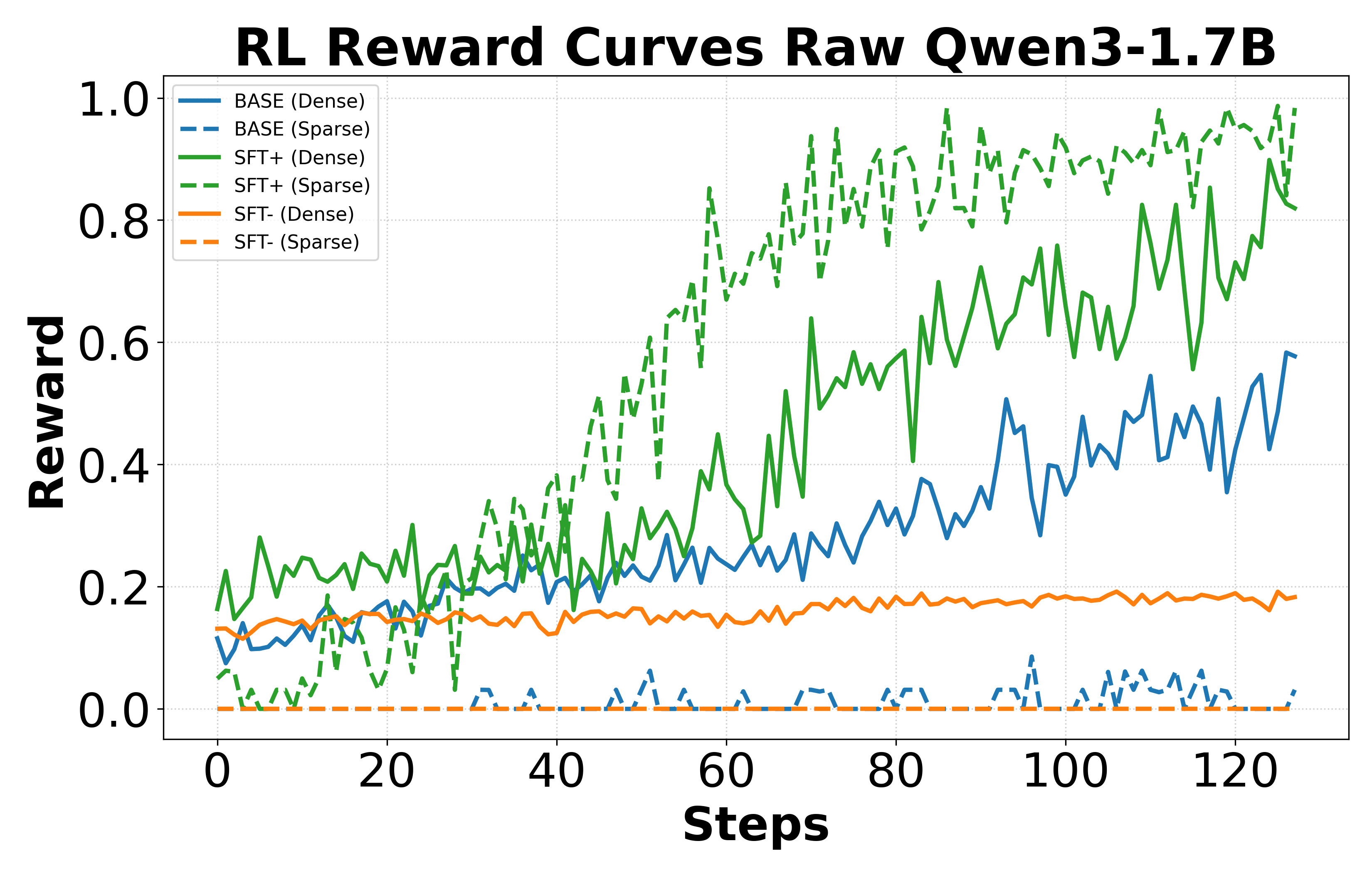}
        \caption{Qwen3-1.7B, movie quote}
        \label{fig:dense_overlay_1b}
    \end{subfigure}
    \hfill
    \begin{subfigure}{0.49\linewidth}
        \includegraphics[width=\linewidth]{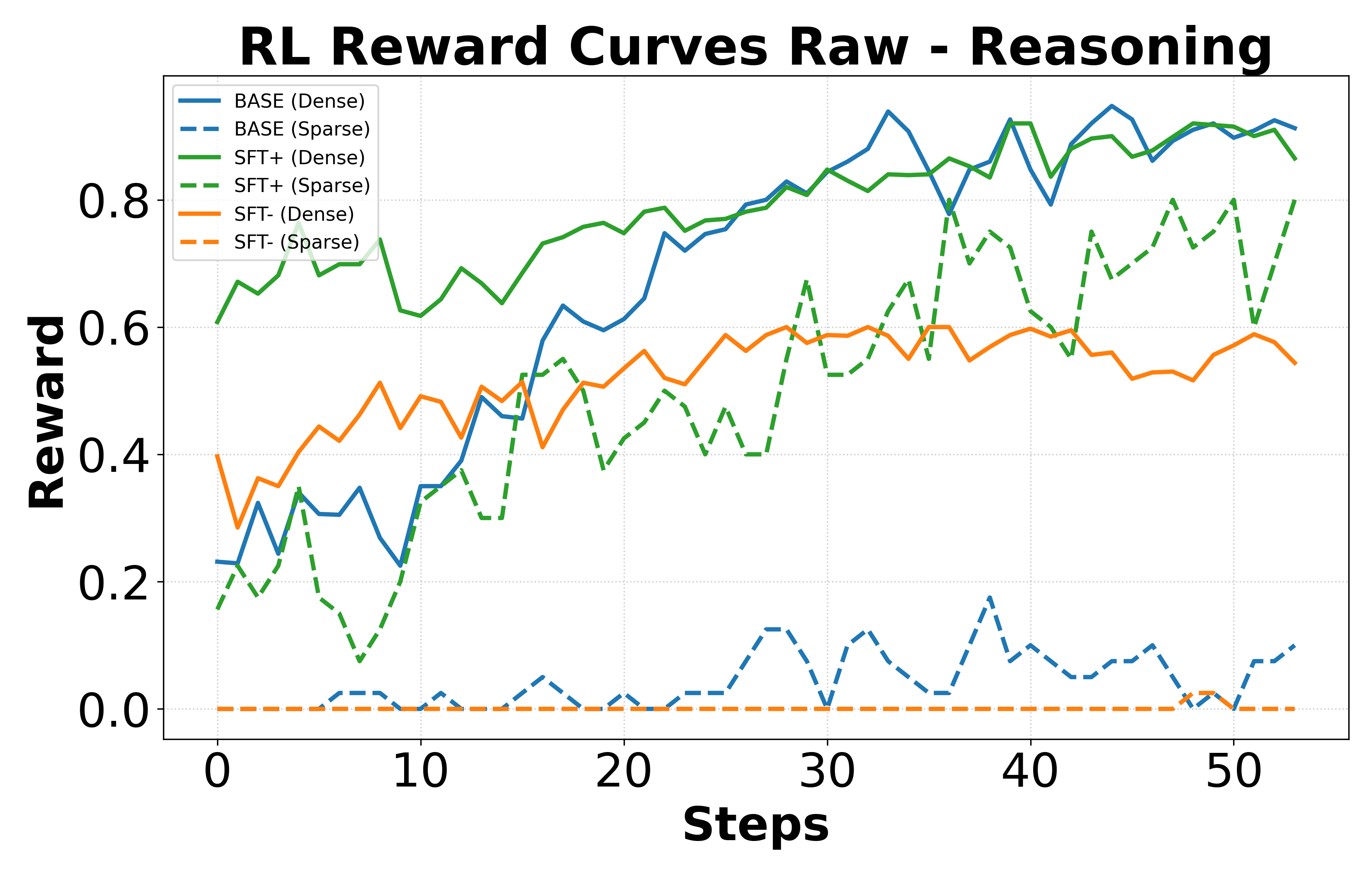}
        \caption{Qwen2.5-7B-Instruct, AIME}
        \label{fig:dense_overlay_7b}
    \end{subfigure}
    \caption{Overlay of sparse vs.\ dense learning dynamics. RL learning curves for \textbf{(a)} Qwen3-1.7B on the movie quote task; 
    \textbf{(b)} Qwen2.5-7B-Instruct on the AIME task. Both figures show an example of how dense rewards can successfully guide the model to learn a behavior with very low probability in the base model, which it is not possible to learn with sparse rewards.
   }
    \label{fig:dense_overlay}
\end{figure*}

A prevailing assumption in recent alignment literature is that RL post-training cannot formulate new behaviors, but rather strictly upweights the existing ``pass@k'' capabilities of the base model, acting merely as a selector rather than a driver of novel behavior \citep{yue2025doesreinforcementlearningreally}. Our next experiment challenges this assumption, by testing if providing the model with sufficiently dense yet accurate rewards can enable learning new behaviors.

\textbf{Target string generation.} We first experiment in our sequence-generation sandbox, by replacing the binary success metric with a dense Levenshtein distance reward to provide the model with intermediate gradient signals based on structural proximity to the target. Figure \ref{fig:dense_overlay_1b} shows learning curves for the 
Qwen3-1.7B model trained with both sparse and dense rewards. Despite the base model possessing a marginal $\approx 0.5\%$ base probability of producing $\tau$, with post-training failing under standard sparse rewards, the dense signal successfully guided the model out of failure, pulling its final match rate to nearly 50\%. However, as in classical RL we find that reward shaping has its limits; the SFT+ model trained with dense rewards underperforms the sparse variant in absolute string matching performance, likely because the Levenshtein distance does not sufficiently incentivize producing the \textit{exact} target string over close approximations.

\textbf{RLVR reasoning.} We then assess if these results replicate for more complex reasoning tasks, testing whether dense rewards provided through a PRM can overcome base model exploration bottlenecks on an AIME mathematical problem. As Figure \ref{fig:dense_overlay_7b} shows, when optimizing with standard sparse rewards on this task, the base reasoning model failed to meaningfully improve, flat-lining at a match rate of approximately 10\% (see Table \ref{tab:pre-training-dist-aime}). 
By contrast, introducing the dense reward radically altered the learning dynamics. With exponentially increasing dense rewards for correctly reaching each step of the reasoning problem, the Base model's exact match rate jumped to over 90\%. It even slightly outperformed the artificially inflated SFT+ model, which plateaued around 85\% under both reward types. Qualitative analysis of the Base model's generated trajectories reveals that the dense reward successfully guided the model to combine disparate, partial reasoning steps into novel, valid solution paths that differed from the original SFT+ ground truth. This shows that RLVR can achieve complex generalization, provided the reward signal is granular enough to credit intermediate logic.

The SFT- reasoning experiments also highlight an interesting nuance about partial learning. While the heavily penalized SFT- model achieved a 0\% exact match rate under both sparse and dense rewards, tracking its trajectory revealed that under the dense reward, its average reward climbed steadily to approximately 0.6 (Figure \ref{fig:dense_overlay_7b}). This indicates that even when a policy is actively suppressed from reaching the final correct answer, a dense reward allows it to re-learn through iteratively building on intermediate reasoning milestones.

Taken together, these results demonstrate that the limitations often attributed to RLVR---such as the inability to move beyond pass@k priors---are not inherent flaws of RL post-training, but rather artifacts of the exploration bottleneck caused by sparse, binary reward functions.

\subsection{Prompt Distribution: Broad vs.\ Narrow}
\label{sec:prompt_dist_results}

\begin{figure}[t]
    \centering
    \includegraphics[width=\columnwidth]{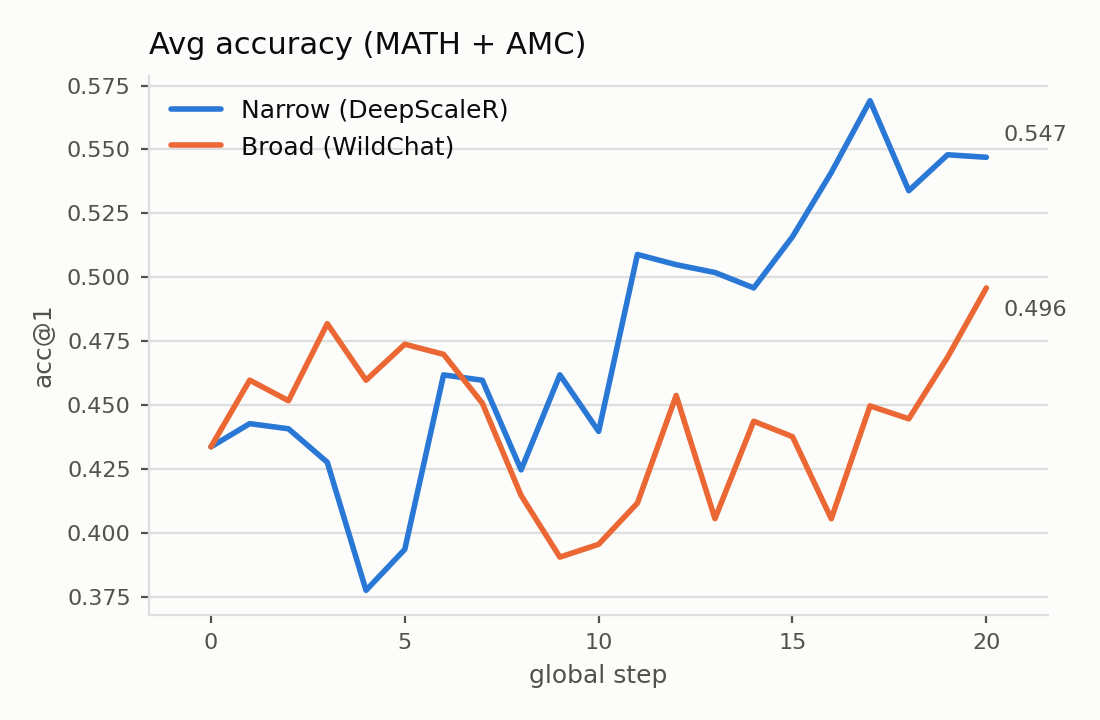}
    \caption{Acc@1 on the MATH and AMC datasets after training on narrow and broad $\mathcal{D}$; training on narrow $\mathcal{D}$ yields higher accuracy.}
    \label{fig:prompt_distribution}
\end{figure}

\begin{figure*}[t]
    \centering
    \begin{subfigure}{0.32\linewidth}
        \includegraphics[width=\linewidth]{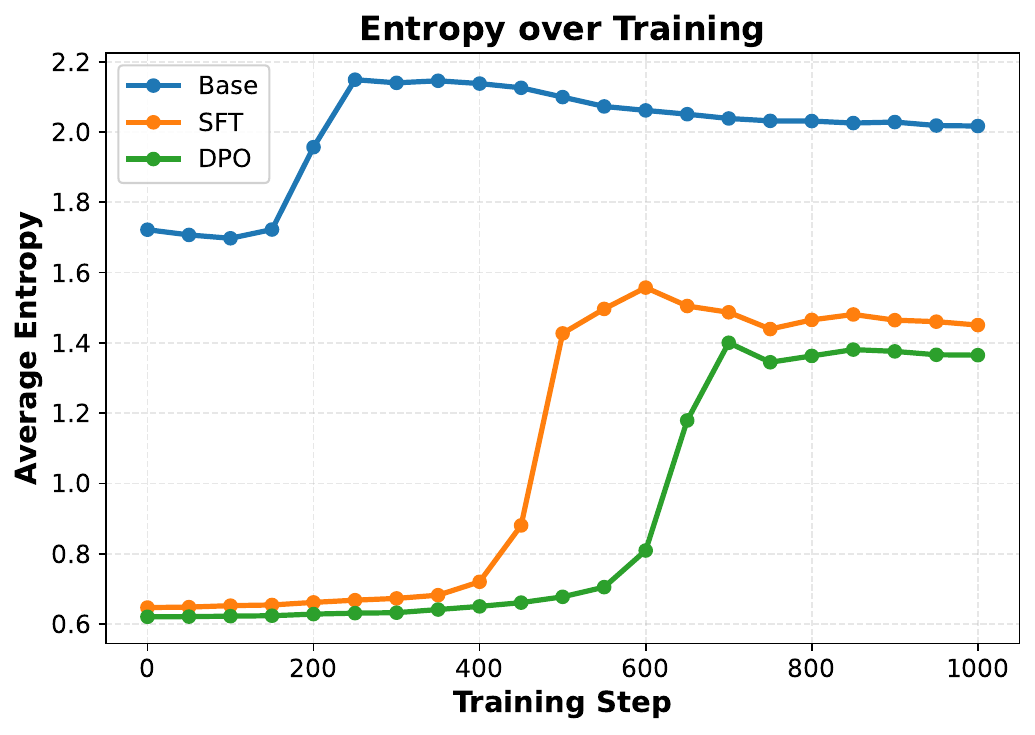}
        \caption{Broad distribution: entropy}
        \label{fig:entropy_all}
    \end{subfigure}
    \hfill
    \begin{subfigure}{0.32\linewidth}
        \includegraphics[width=\linewidth]{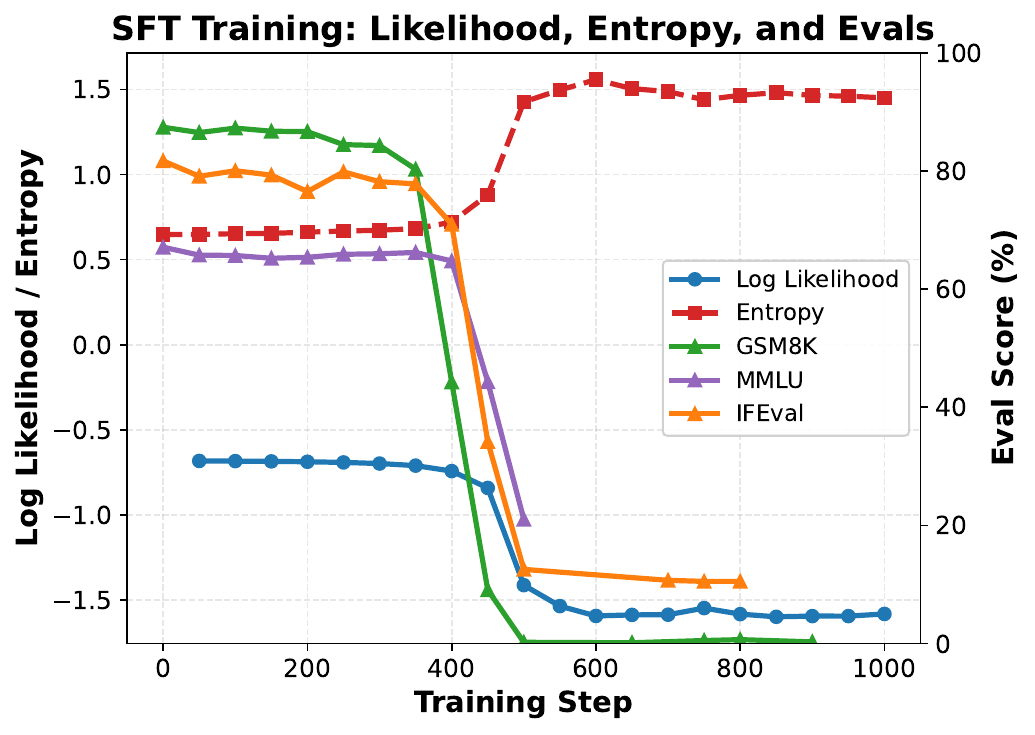}
        \caption{Broad distribution: capabilities}
        \label{fig:entropy_broad}
    \end{subfigure}
    \hfill
    \begin{subfigure}{0.32\linewidth}
        \includegraphics[width=\linewidth]{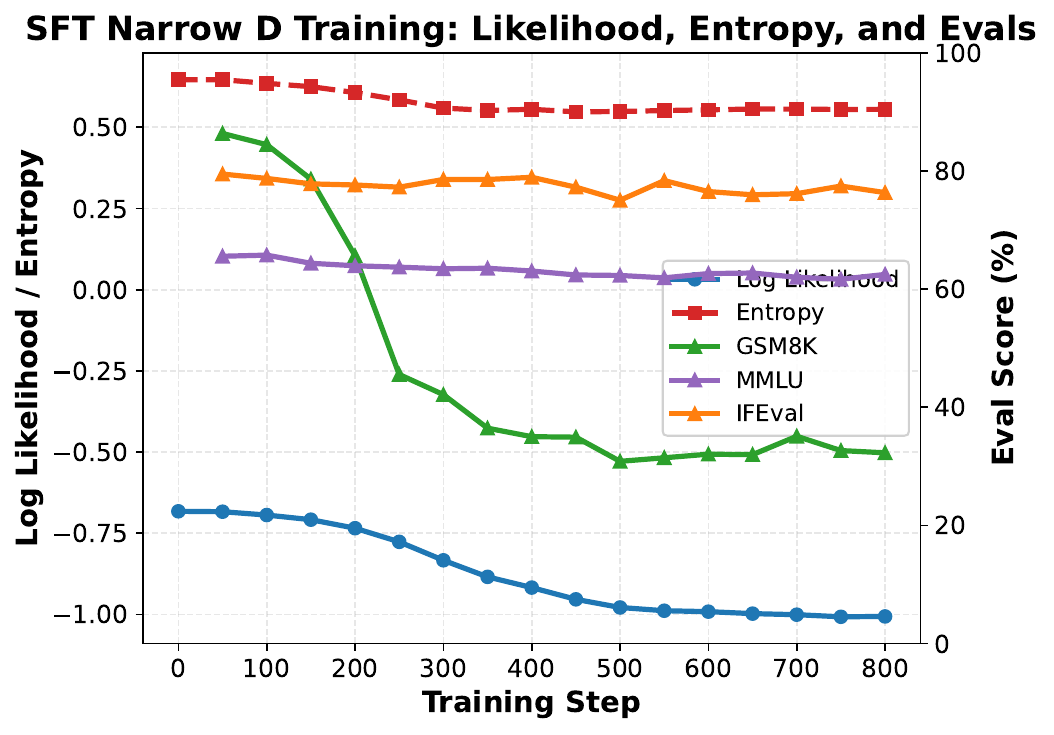}
        \caption{Narrow distribution}
        \label{fig:entropy_narrow}
    \end{subfigure}
    \caption{Effect of training OLMo models on spurious (random) rewards). \textbf{(a)} Average entropy on broad $\mathcal{D}$ (10k prompts). Each training stage (Base $\to$ SFT $\to$ DPO) delays entropy increase onset, showing increasing resilience. \textbf{(b)} Broad $\mathcal{D}$ SFT model---entropy spike at step $\sim$400 coincides with collapse of all eval metrics (GSM8K, MMLU, IFEval), confirming global unlearning. \textbf{(c)} Narrow $\mathcal{D}$ SFT model (100 math prompts)---entropy \emph{decreases}, MMLU and IFEval are preserved, but GSM8K collapses, showing targeted corruption of the training domain.}
    \label{fig:new_prompt_distribution}
\end{figure*}

\textbf{Qwen family.} Figure~\ref{fig:prompt_distribution} show the effects of training on spurious (random) rewards, with both the narrow prompt distribution of \citet{shao2025spuriousrewardsrethinkingtraining}, and the broad distribution. The narrow distribution replicates the original results, showing that the capabilities of Qwen models can actually increase with random rewards. However, when switching to a broad prompt distribution, we see minimal improvements on the MATH benchmark, and degraded performance on AMC. This is supported by tracking the entropy through training (Figure~\ref{fig:new_prompt_distribution}). Under the narrow distribution, entropy remains low through training, while under the broad distribution, it jumps at the first step and remains high. 

\textbf{OLMo family.} Figure \ref{fig:new_prompt_distribution} shows the effect of post-training the OLMo family of models (base, SFT, and DPO) on spurious rewards.
Under a broad prompt distribution $\mathcal{D}$, we see that the entropy of the models increases (Fig. \ref{fig:entropy_all}), and that this increase at step $\sim$400 for the SFT model corresponds to a sharp decrease in capabilities, as measured with the 
GSM8K, MMLU, and IFEval benchmarks (Fig. \ref{fig:entropy_broad}).   
Notably, log likelihood decreases and entropy increases uniformly across all three domains (math, code, instruction following), with no generalization gap. This indicates the unlearning is global: random rewards on a broad $\mathcal{D}$ effectively reward any response to any prompt, erasing the model's learned distribution.

Under a narrow $\mathcal{D}$ (100 math-only prompts), the dynamics differ strikingly. As shown in Figure~\ref{fig:entropy_narrow}, entropy \emph{decreases} slightly across all domains, indicating the model's distribution is tightening rather than dispersing. However, this increased certainty does not mean improved performance: GSM8K drops from $86\%$ to $\sim$32\% by step 400, while MMLU ($\sim$65\%$\to$62\%) and IFEval ($\sim$79\%$\to$77\%) remain largely intact. Damage is domain-specific: random rewards corrupt the training domain while preserving capabilities in unsampled domains.

The breadth of $\mathcal{D}$, as well as the base model's distribution, determine both the scope of degradation, and whether degradation occurs. A broad $\mathcal{D}$ causes widespread unlearning. A narrow $\mathcal{D}$ increases certainty about the responses the base model was already likely to produce, which can either lead to performance improvement or targeted corruption, depending on the base model's initial accuracy. Although a narrow $\mathcal{D}$ uses fewer examples with more repetition, this repetition is exactly what drives the policy to collapse toward its prior. Neither setting creates new capabilities. Qwen's gains under a narrow $\mathcal{D}$ fit this explanation: the same targeted sharpening that harms GSM8K performance in OLMo 3 instead reinforces the correct math reasoning paths already present in Qwen. This suggests that spurious rewards improve performance only when the prompt distribution is narrow and the base model is already strongly biased toward the target domain, a restricted setting that does not apply to most of RL post-training.

\section{Conclusion}
In this work, we deconstructed the mechanics of RL post-training for LLMs to explain how components of the post-training algorithm---the base model distribution, prompt distribution, and reward structure---affect optimization outcomes. By isolating the components of the RLVR pipeline within a controlled sequence-generation sandbox, we explain, empirically verify, or provide a counterpoint to existing post-training results in the literature. In particular, we show that standard sparse RL is unable to surpass the base model's initialization limits. When a target behavior lacks sufficient baseline coverage, the model cannot sample it, making optimization untenable.
However, we show that this is not a hard and fast limitation of RL post-training as a technique; a sufficiently dense and accurate reward function can enable learning new behaviors with vanishing support in the base model, a finding that contradicts a popular position in the literature \cite{yue2025doesreinforcementlearningreally,shao2025spuriousrewardsrethinkingtraining,wu2507invisible,zhao2025echo,wu2025position}. 
Furthermore, we show the effect of ``spurious rewards'' depends entirely on the post-training prompt distribution. Training with random rewards on a broad prompt distribution leads to an increase in model entropy and a corresponding catastrophic decrease in capabilities.
By clarifying these interacting mechanics, we hope this work can serve as a resource to the NLP community and those looking to sharpen their understanding of the mechanics of RL post-training, and help move the field beyond treating RL as a ``black box'' toward a more controlled and interpretable optimization process.

\section*{Acknowledgments}
This research was supported by the UW-Amazon Science Gift Hub, UW-Tsukuba Amazon NVIDIA Cross Pacific AI Initiative (XPAI), Sony Research Award, Tinker Research Grants, Character.AI, DoorDash, Open Philanthropy, Coefficient Giving, Toyota Research Institute, the Schmidt AI2050 Fellows program, and the NSF CISE RI program,  award \#2550849. This material is based upon work supported by the Defense Advanced Research Projects Agency and the Air Force Research Laboratory, contract number(s): FA8650-23-C-7316. Any opinions, findings and conclusions, or recommendations expressed in this material are those of the author(s) and do not necessarily reflect the views of AFRL or DARPA. This work was supported by NSF grants CNS-2112471, IIS-2229876, CCF-2505865, and DMS-2502281.

\bibliography{acl_natbib}

\appendix
\section{Resource Details}

\subsection{Computational Specifications}
All experiments were run on either NVIDIA A100 80GB GPU or NVIDIA L40s 48GB. For the Math reasoning task, an NVIDIA A100 80GB GPU was used for the main experiments, while an NVIDIA L40s 48GB was used for the PRM LLM-as-a-Judge.
\subsection{LLM Usage}
LLMs were used in the development of the code base/experiments, specifically modifying behavior to adapt to resource limitations, as well as simplify experiment pipeline for generating results. 

% \newpage
\section{Expanded Related Work}
\label{app:expanded_related_work}

Post-training with RL has proven effective for a wide range of tasks, particularly for improving reasoning capabilities and aligning LLMs with human preferences. For instance, RL with Verifiable Rewards (RLVR) has shown that models can achieve state-of-the-art performance with smaller parameter counts, such as DeepSeek-R1 \citep{Guo_2025, lambert2025tulu3pushingfrontiers}, highlighting its role in developing efficient AI. Through Reinforcement Learning from Human Feedback (RLHF), reward signals are learned from preference data, leading to improvements in instruction following and response quality \citep{ouyang2022traininglanguagemodelsfollow, bai2022constitutionalaiharmlessnessai, stiennon2022learningsummarizehumanfeedback}. Across both settings, RL post-training has shown to yield strong gains on reasoning and alignment tasks, making a mechanistic understanding of its success essential.

Several recent works have surfaced seemingly contradictory findings about RL post-training. \citet{shao2025spuriousrewardsrethinkingtraining} demonstrate that RLVR can yield substantial gains in mathematical reasoning, even when reward signals are weak, random, or negatively correlated with the actual answer. Notably, they find that spurious rewards still trigger systematic behavioral shifts, such as increased code reasoning frequency in Qwen2.5-Math, suggesting the base model's pretraining dictates the form of emergent behavior, not the reward. Moreover, they observe that these gains do not generalize consistently across different model families; for example, Qwen2.5 and OLMo2 exhibit different behavioral trends to these spurious rewards. These findings appear to challenge a core idea in RL: that improvements arise from optimizing correctly specified reward signals. Our work addresses these questions by showing that the effect of random rewards depends critically on the prompt distribution. A broad prompt distribution causes global entropy increase and capability degradation, while a narrow distribution causes targeted corruption of the training domain while preserving other capabilities. The prompt distribution thus controls the \textit{scope} of degradation, not whether degradation occurs.

Moreover, the original base model plays a central role in determining post-training success \citep{yue2025doesreinforcementlearningreally}. Measuring pass@k (the probability of solving a problem in $k$ attempts), the authors find that while RL-trained models excel with small values of $k$, base models catch up as $k$ increases. This suggests RLVR primarily improves sampling efficiency, refining existing behaviors rather than creating novel ones. We build on this by comparing outcomes across models with different baseline capability distributions under controlled reward conditions, and show that a sufficiently shaped reward can overcome this problem, and teach new behaviors to models that previously did not generate them with high probability. 

\citet{chen2025coverageprinciplepretrainingenables} introduce the ``Coverage Principle`` and argue that existing metrics (e.g. cross-entropy loss) are often poor predictors of post-training success, instead stating the coverage (the total probability mass the pre-trained model assigns to the set of high-quality responses) is the critical link. This motivates our choice to treat this notion of coverage as an independent variable; by selecting base models with varying coverage over target tasks, we can test how it interacts with the effect of reward density, allowing us to disentangle these two sources of variation.

\citet{ren2025learningdynamicsllmfinetuning} develop a framework to illustrate why post-training often redistributes probability mass rather than creating new capabilities by decomposing changes in log-probabilities after gradient updates. While their results are restricted to Direct Preference Optimization (DPO), this perspective accounts for observed behavioral shifts such as hallucinations and highlights the complex dynamics underlying alignment and performance gains. We adopt this framing for interpreting behavioral changes, as gains in the target task likely reflect probability mass redistribution rather than the emergence of new capabilities.

\section{Reward Function Implementations - Movie Quotes}
\label{app:reward_implementations}

In our sandbox experiments, we utilize two distinct reward functions to evaluate the impact of reward shaping on RL optimization: a sparse reward (equipped with an anti-rambling penalty) and a dense Levenshtein-based reward. Let $y$ denote the model's generated sequence and $\tau$ denote the exact target string.

\subsection{Sparse Reward with Length Penalty}
The sparse reward acts primarily as a binary success metric. It assigns a base reward of 1 if the target string $\tau$ is successfully generated, and 0 if the model fails to produce the target entirely.

To prevent reward hacking by ``rambling'', we use a length penalty term $p$ to decrease the reward for any extra tokens generated other than the target string. Let $n$ be the maximum token generation limit and $s$ be the number of extra tokens generated. Then, $p$ is defined as
\begin{equation}
    p = \begin{cases}
        0 & \text{if $s = 0$} \\
        \frac{s}{n} & \text{if $s > 0$}
    \end{cases}
\end{equation}
and the sparse reward function is defined as
\begin{equation}
    r_{\text{sparse}}(y, \tau) = 
    \begin{cases} 
    \max(0.5, 1 - p) & \text{if } \tau \in y \text{ and } |y| > |\tau| \\
    0 & \text{otherwise} 
    \end{cases}
    \label{spare_rewards}
\end{equation}
*(Note: $p$ can be defined as a fixed scalar penalty or a function of the excess length $(|y| - |\tau|)$).* This ensures the sparse optimization baseline actively discourages rambling and aligns only the precise target behavior.

\subsection{Dense Reward (Levenshtein Distance)}
To address the credit assignment problem for models lacking initial probability mass, we implement a dense reward that provides continuous, intermediate gradient signals based on structural proximity to the target. 

We utilize the standard Levenshtein edit distance, which calculates the minimum number of single-character edits (insertions, deletions, or substitutions) required to transform the generated sequence into the target sequence. Formally, let $y$ and $\tau$ be strings of lengths $m$ and $n$, and let $D(i,j)$ denote the Levenshtein distance between the first $i$ characters of $y$ and $j$ characters of $\tau$, where $y_i$ and $\tau_j$ are the $i$-th and $j$-th characters. The Levenshtein distance is defined for boundary cases as $D(i, j) = \max(i, j)$ when $i=0$ or $j=0$. Otherwise, it is calculated as:
\begin{equation}
D(i, j) = \min \begin{cases}
D(i-1, j) + 1\\
D(i, j-1) + 1\\
D(i-1, j-1) + \mathbb{1}(y_i \neq \tau_j)
\end{cases}
\label{lev_dist}
\end{equation}

Let $n = \max(\text{len}(y), \text{len}(\tau))$ and $D$ be the Levenshtein distance between $y$ and $\tau$. Then, reward function is then defined as 
\begin{equation}
    r_{dense} = \max(0, 1 - \frac{D}{n})
\end{equation}

\section{Exponential Process Reward Model (PRM) Implementation}
\label{app:prm_details}

For our math reasoning experiment, in order to scale our analysis from structural sequence-matching (Levenshtein distance) to combinatorial mathematical reasoning, we replaced the string-based dense reward with a Process Reward Model (PRM). Mathematical reasoning requires logical verification across diverse, valid solution paths (e.g., direct factoring, quadratic formula, de-homogenization). To achieve this, we deployed a 32-billion parameter instruction-tuned model (Qwen2.5-32B-Instruct) functioning as an LLM-as-a-Judge. 

The PRM evaluates the generated trajectory $y$ at the end of each rollout, classifying the strategy used and evaluating the presence of specific, sequential logical milestones.

\subsection{Logical Milestones and Reward Curve}
The AIME problem utilized in our experiments ($12x^2 - xy - 6y^2 = 0$) requires a multi-step algebraic derivation. We instruct the PRM to extract a boolean vector $M(y) \in \{0, 1\}^5$, representing the successful completion of the following five milestones:
\begin{itemize}
    \item $m_1$: Initiated a correct algebraic path configuration (e.g., setting up factorization).
    \item $m_2$: Derived accurate base root relationships between $x$ and $y$.
    \item $m_3$: Evaluated the integer boundary constraints ($-100 \le x, y \le 100$) correctly.
    \item $m_4$: Calculated the correct solution count for at least one case branch.
    \item $m_5$: Identified and correctly subtracted the intersection overlap at the origin $(0,0)$.
\end{itemize}

A critical challenge in applying linear partial credit (e.g., $+0.15$ per step) is the introduction of local minima. An optimizing policy may discover that completing three steps and halting yields a ``good enough'' reward, dampening the advantage gradient required to risk generating further tokens to reach the final answer. To prevent this advantage compression, we map the milestones to a reward curve. The step weights are defined as $w = [0.05, 0.05, 0.10, 0.15, 0.25]$. The base process reward is calculated as:
\begin{equation}
    r_{process}(y) = \sum_{i=1}^{5} w_i \cdot m_i
\end{equation}
This exponential scaling strictly enforces increasing marginal value, ensuring that advanced reasoning steps provide a stronger gradient pull than early exploratory steps.

\subsection{Outcome Overrides and Defensive Penalties}
To anchor the PRM to the absolute ground truth and prevent reward hacking, the final reward incorporates deterministic outcome overrides and active penalties. 

Let $\tau = 117$ be the exact correct answer, and $\tau_{near} = 118$ represent the specific, known near-miss where the model successfully completes the calculus but fails to account for the origin overlap. Let $\text{extract}(y)$ be a deterministic parsing function that extracts the integer from the final \verb|\boxed{}| command. The unpenalized reward $r_{base}$ is defined as:

\begin{equation}
r_{base}(y) = 
\begin{cases} 
1.0 & \text{if } \text{extract}(y) = \tau \\
0.6 & \text{if } \text{extract}(y) = \tau_{near} \\
r_{process}(y) & \text{otherwise}
\end{cases}
\end{equation}

Finally, LLM policies optimizing against dense rewards frequently attempt to ``farm'' tokens by redundantly repeating early logical steps that are known to yield partial credit. To counteract this, the PRM is instructed to flag cyclic reasoning with a boolean penalty marker $m_{loop}$. If the model loops without reaching the final exact answer, a penalty $p_{loop} = -0.3$ is applied. Furthermore, a strict formatting penalty $p_{format} = -0.5$ is applied if the trajectory fails to conclude with the \verb|\boxed{}| delimiter. 

The final dense reward is bounded at $0$ and defined as:
\begin{equation}
    r_{dense}(y) = \max\Big(0, \; r_{base}(y) + p_{loop} + p_{format}\Big)
\end{equation}

By heavily penalizing redundancies and exponentially rewarding progress, this PRM structure successfully bridges the exploration gap for models with near-zero initial priors.

\section{RLVR Algorithm}

\begin{algorithm}[H]
\caption{RLVR Algorithm}
\label{alg:RL-algo}
    \begin{algorithmic}[1]
       \STATE {\bfseries Input:} base model $\pi_{ref}$, prompt dataset $D$, reward function $r(x, y)$
       \STATE {\bfseries Initialize:} $\pi_\theta \leftarrow \pi_{ref}$ \hspace{5pt} %\# initialize policy with base model
       \FOR{train steps}
           \STATE Sample input: $x \sim \mathcal{D}$
           \STATE Sample output: $y \sim \pi_\theta(y \mid x)$
           \STATE  Get reward: $r \gets r(x, y)$
           \STATE Store $\langle x, y, r \rangle$ in buffer
           \IF{Time to train}
                \STATE Update model with an RL algorithm of choice to maximize:\\  \hspace{-0.7cm}{\small $ \mathbb{E}_{ {\rm buffer}} [\frac{1}{c}r(x,y) - D_{\rm KL}( \pi_\theta \| \pi_{\rm ref})]$}
            \ENDIF
       \ENDFOR
       \RETURN{$\pi_\theta$}
\end{algorithmic}
\end{algorithm}

\section{Prompt Distribution Experimental Details}
\label{app:prompt_dist_setup}

\textbf{Training Details:} For the broad-distribution condition detailed in Section \ref{sec:prompt_distributions}, we use a production-scale dataset comprising 10,000 prompts sampled from the OLMo~3 RLVR training mix, evenly split between mathematics, instruction following, and code. We utilize the standard OlmoRL codebase with verifiable rewards replaced by random scalars drawn uniformly from $[0, 1]$. To ensure fully on-policy training, async steps are set to 0. All experiments are distributed across three 8xH100 nodes, sampling 8 responses per prompt. The narrow-distribution condition utilizes the identical hardware and algorithmic setup, but is restricted to 100 math-only prompts fine-tuned directly from the SFT checkpoint. 

\textbf{Evaluation Tracking:} To precisely measure entropy and unlearning dynamics, we extract 200 evaluation prompts per domain (100 that appeared during training, and 100 held-out out-of-distribution prompts). Prior to RL training, we sample a fixed response for each prompt from the base model. Throughout the random reward fine-tuning process, we evaluate the model every 50 training steps, measuring the log-likelihood and average per-token entropy assigned to these fixed baseline responses.

\section{Model Sizes}
\begin{figure}[ht]
    \centering
    \noindent
    \includegraphics[width=0.38\linewidth]{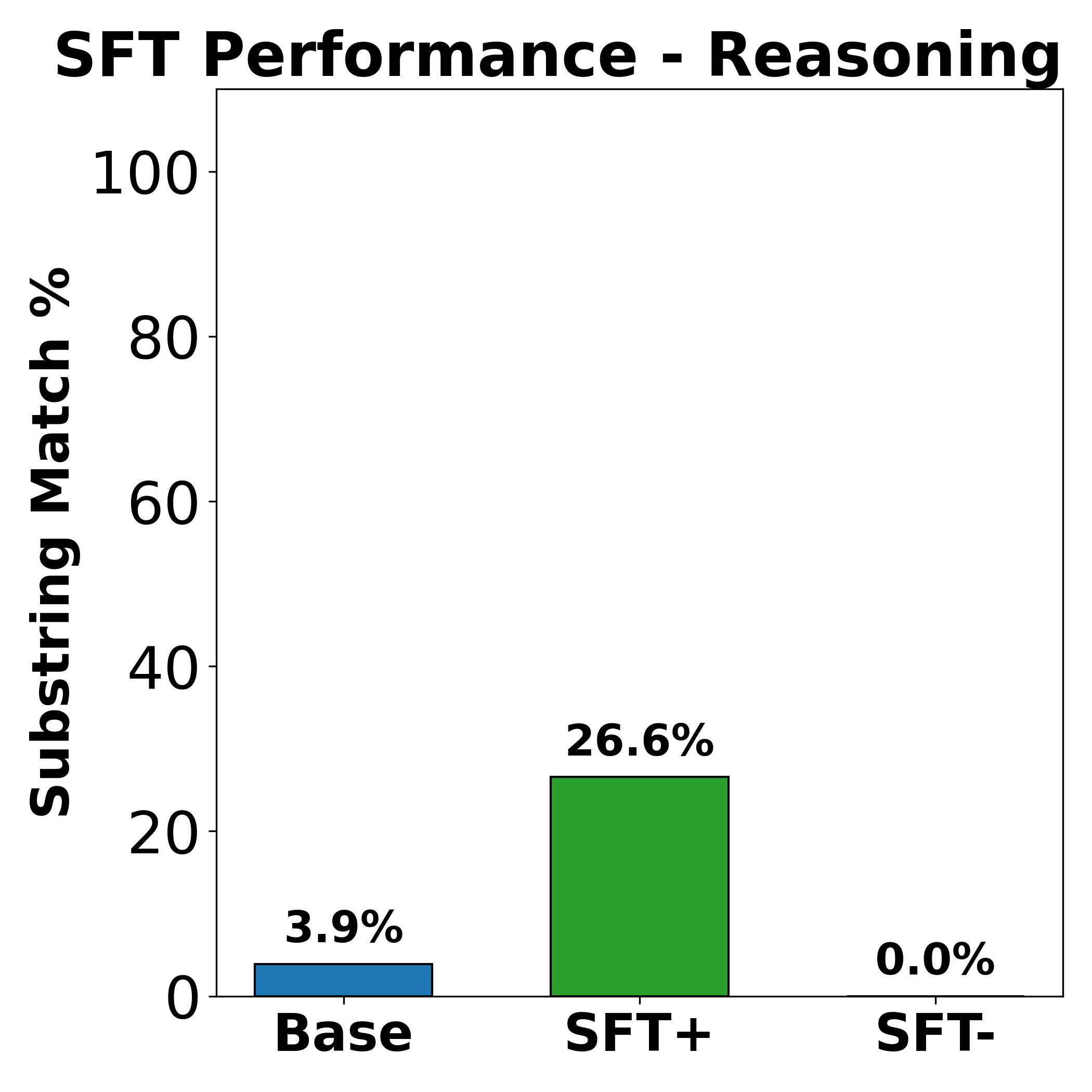}
    \hfill
    \includegraphics[width=0.6\linewidth]{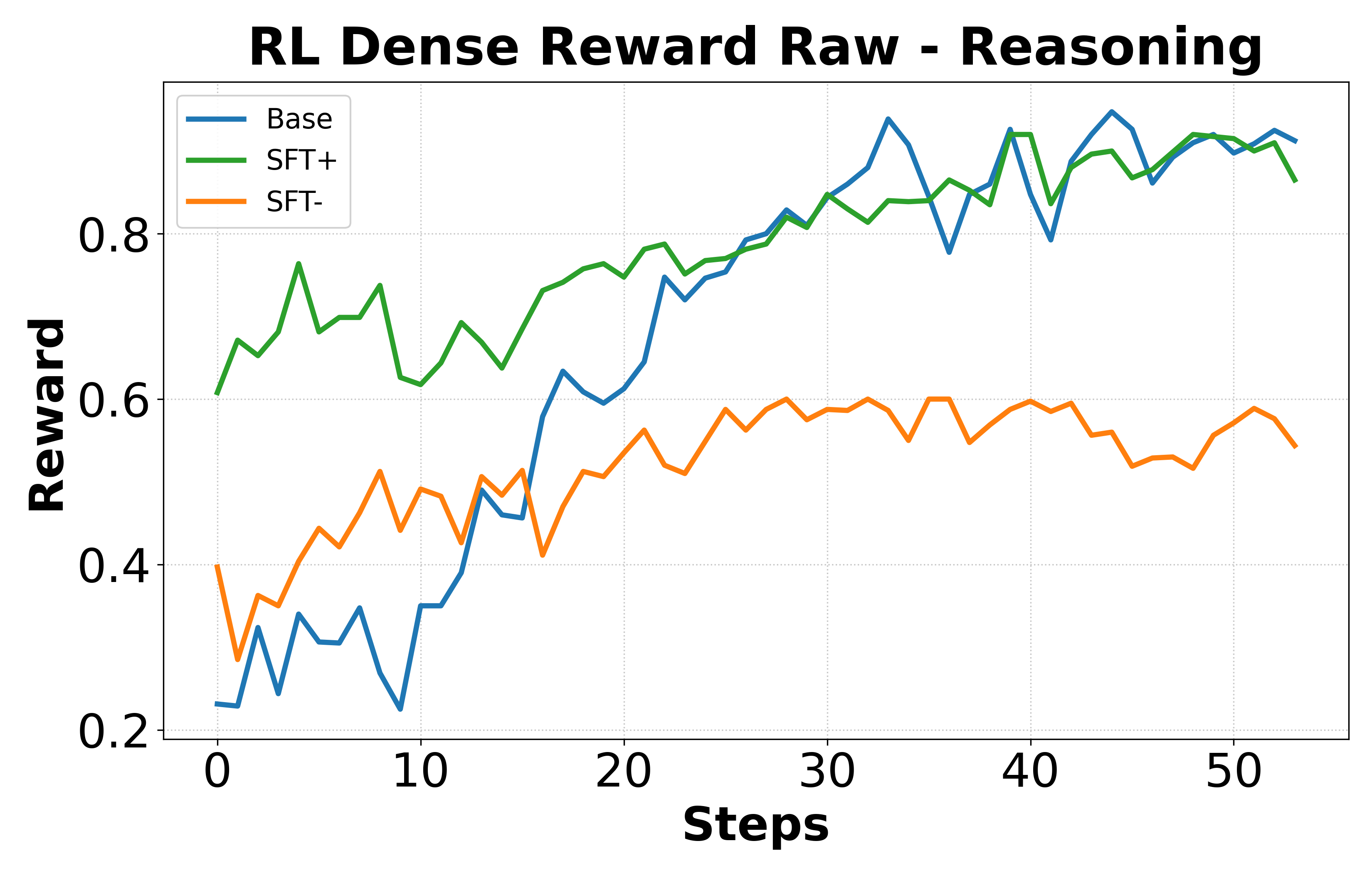}
    \hfill
    \includegraphics[width=0.38\linewidth]{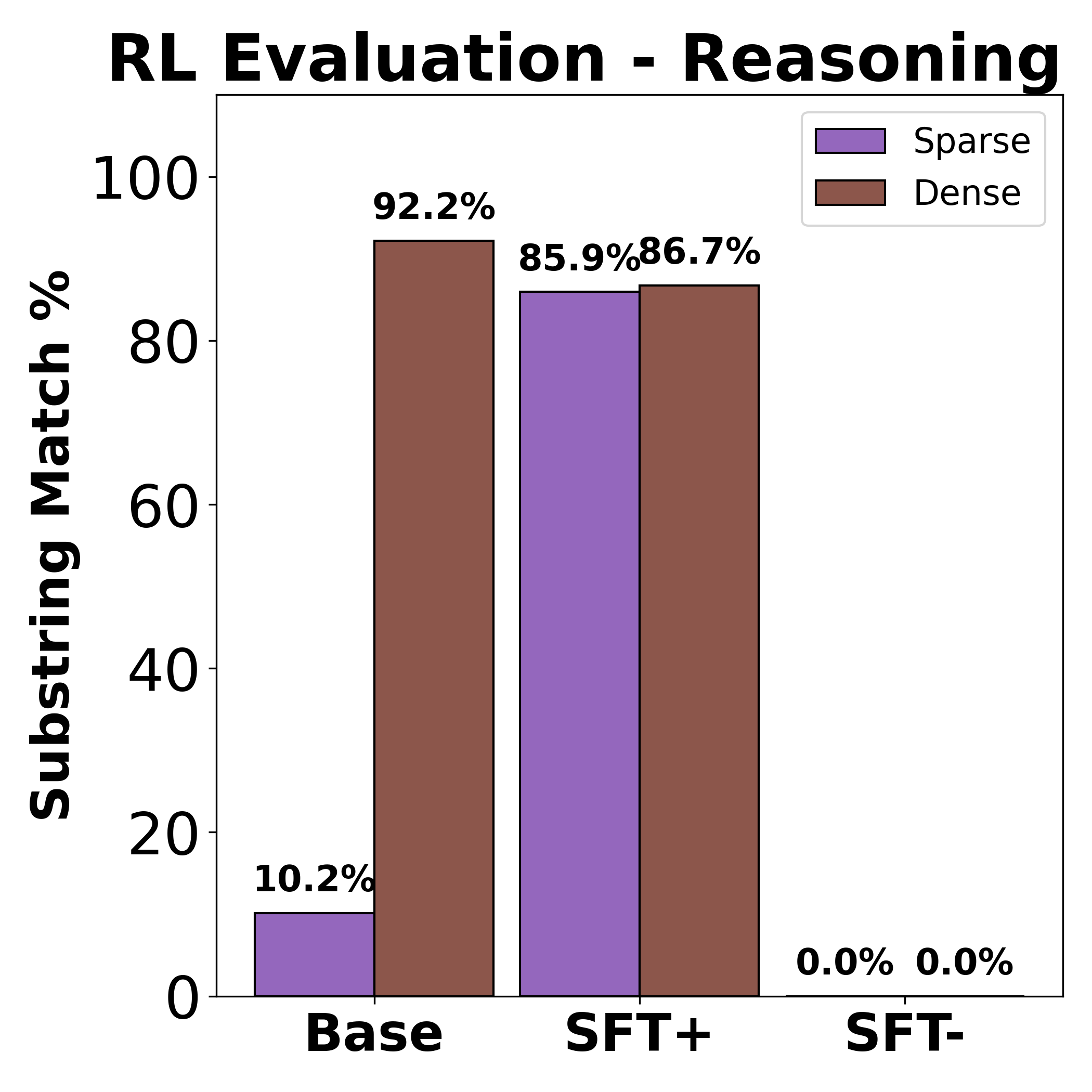}
    \hfill
    \includegraphics[width=0.6\linewidth]{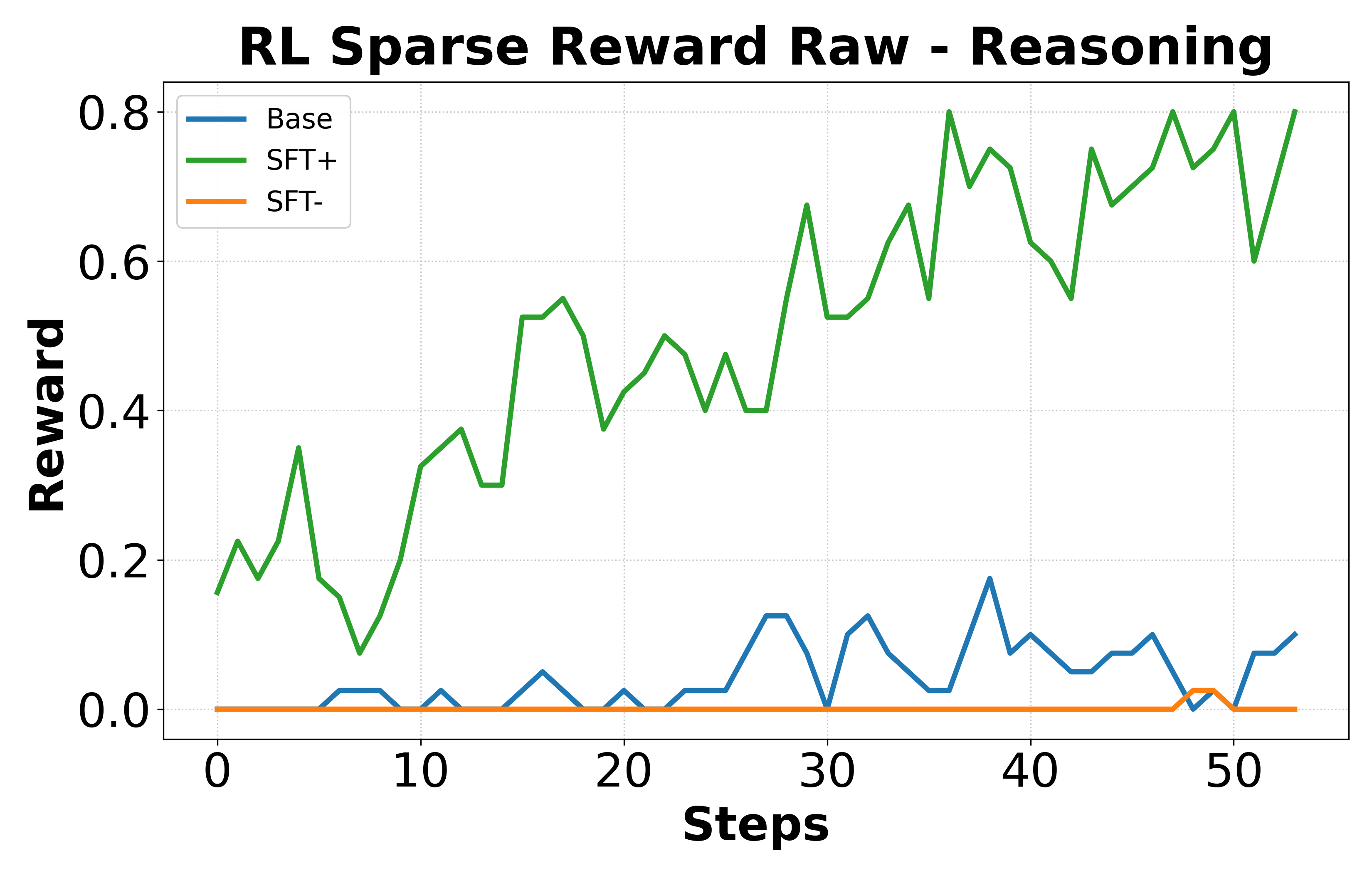}
    \hfill
    \caption{\textbf{RL post-training dynamics and evaluation metrics for Qwen2.5-7B-Instruct - Math Reasoning} \textbf{(Top Left)} Prior to RL, the SFT+ model shows a strong initial prior (26.6\%), and the Base model exhibits a small but non-zero starting mass (3.9\%). \textbf{(Bottom Left)} Final RL evaluation reveals that Dense rewards on both the SFT+ and Base models successfully and acquire the target behavior ($>85\%$ match rates) however under sparse rewards the Base model was unable to acquire the target behavior while the SFT+ model acquires the target behavior. SFT- unable to gain momentum with either reward structure. \textbf{(Right)} The raw reward curves highlight a difference in the RL Training between Base, SFT+ and SFT- models. Experiments show that even through complex reasoning tasks, the priors of the models impact the role of RL post training.}
\end{figure}

\begin{figure}[ht]
    \centering
    \noindent
    \includegraphics[width=0.38\linewidth]{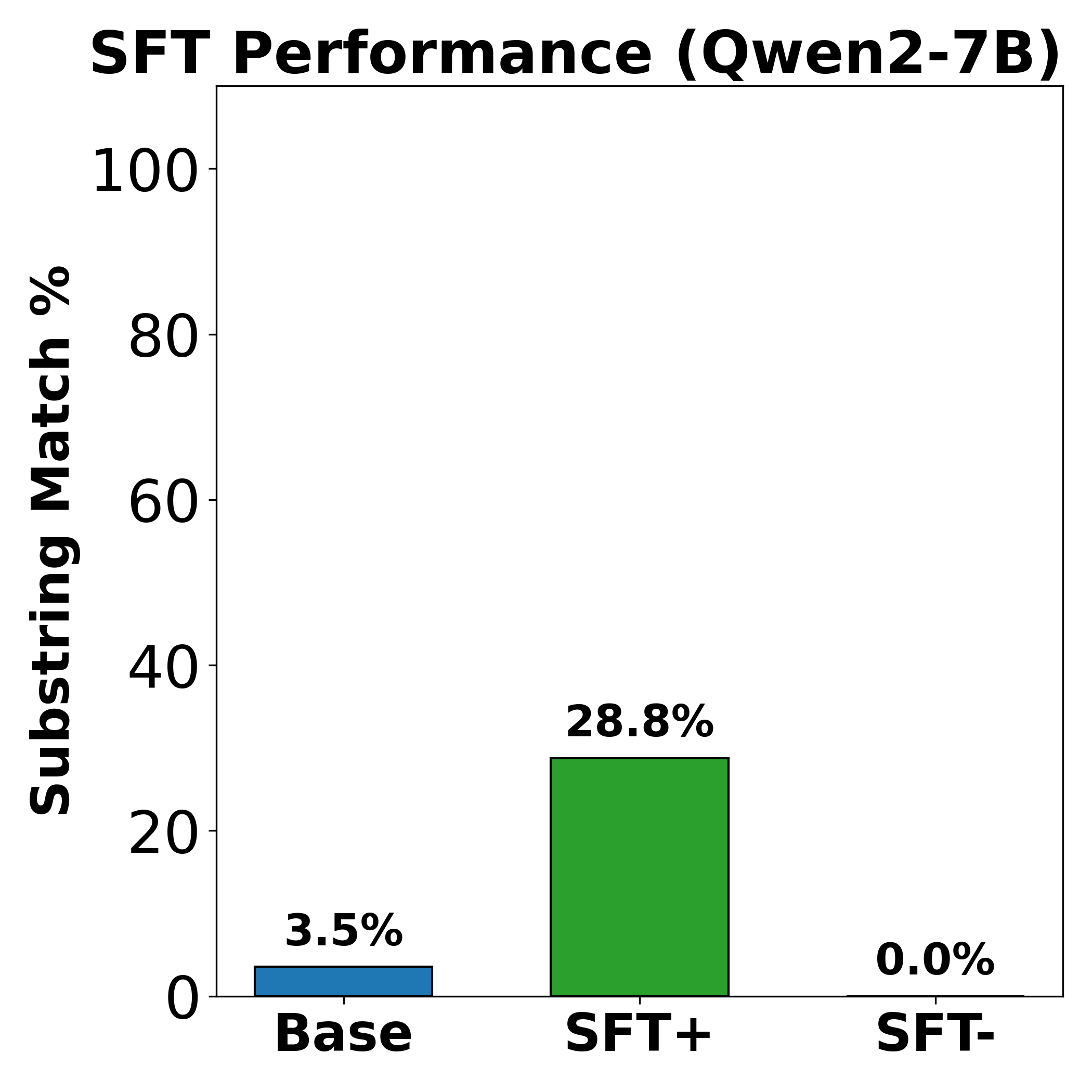}
    \hfill
    \includegraphics[width=0.6\linewidth]{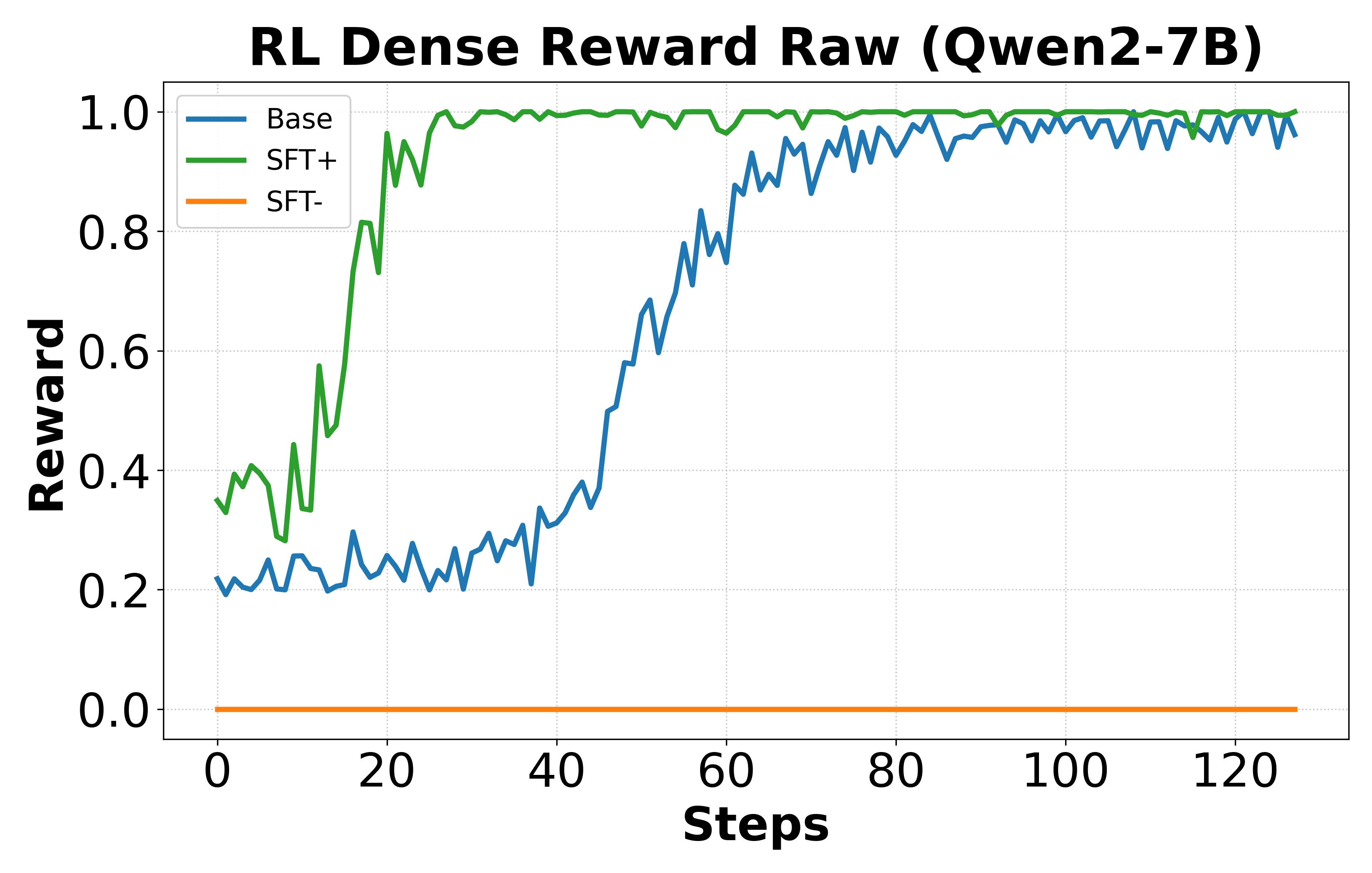}
    \hfill
    \includegraphics[width=0.38\linewidth]{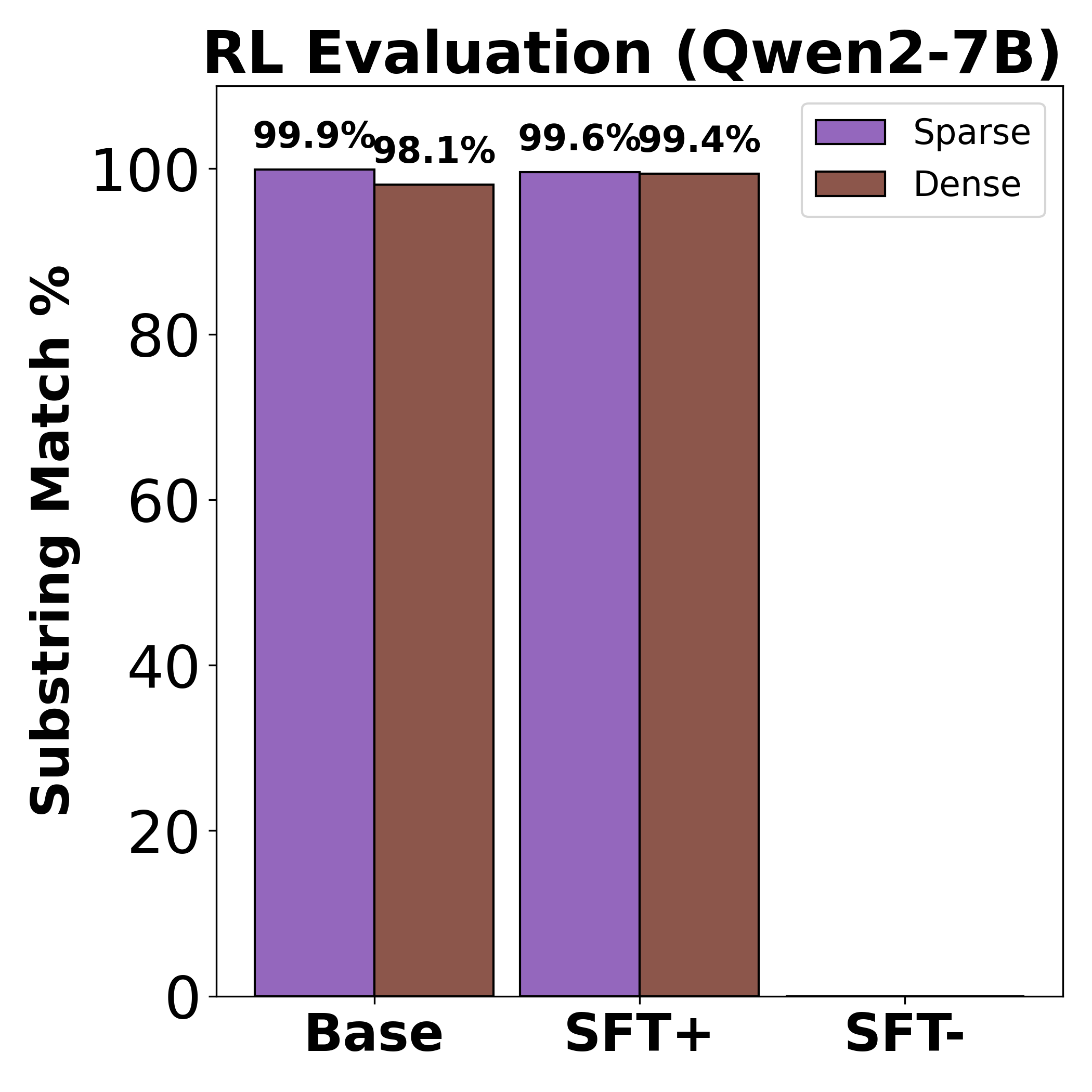}
    \hfill
    \includegraphics[width=0.6\linewidth]{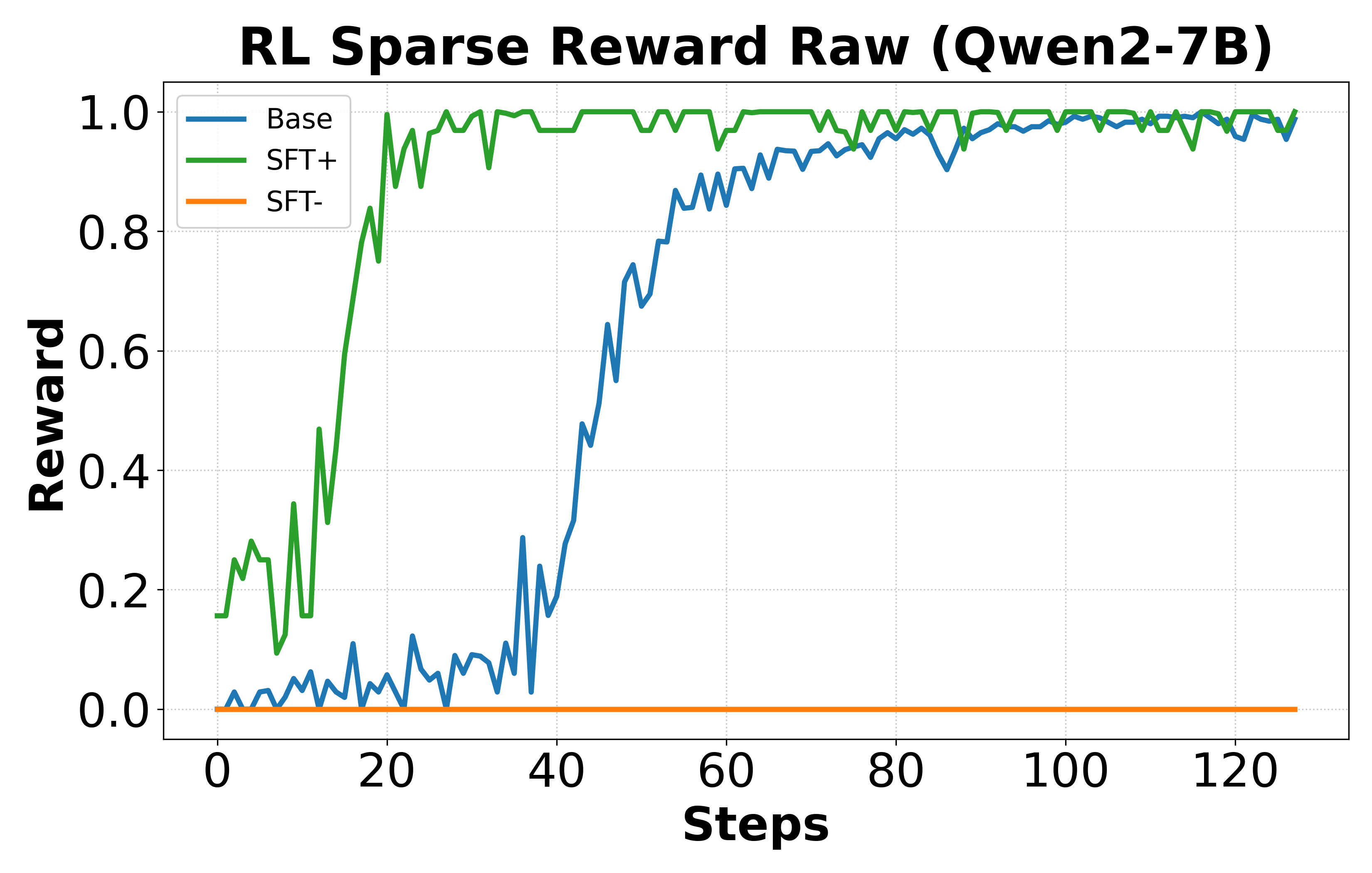}
    \hfill
    \caption{\textbf{RL post-training dynamics and evaluation metrics for Qwen2-7B.} \textbf{(Top Left)} Prior to RL, the SFT+ model shows a strong initial prior (28.8\%), and the Base model exhibits a small but non-zero starting mass (3.5\%). \textbf{(Bottom Left)} Final RL evaluation reveals that both the SFT+ and Base models successfully and perfectly acquire the target behavior ($>98\%$ match rates) under both dense and sparse rewards, while SFT- completely fails. \textbf{(Right)} The raw reward curves highlight a delayed but successful optimization for the Base model. While SFT+ climbs immediately, the Base model requires roughly 40 steps to escape its initial plateau before rapidly converging to a 1.0 reward. The SFT- model remains strictly suppressed.}
    \label{fig:qwen2-7b_Apdx}
\end{figure}

\begin{figure}[ht]
    \centering
    \noindent
    \includegraphics[width=0.38\linewidth]{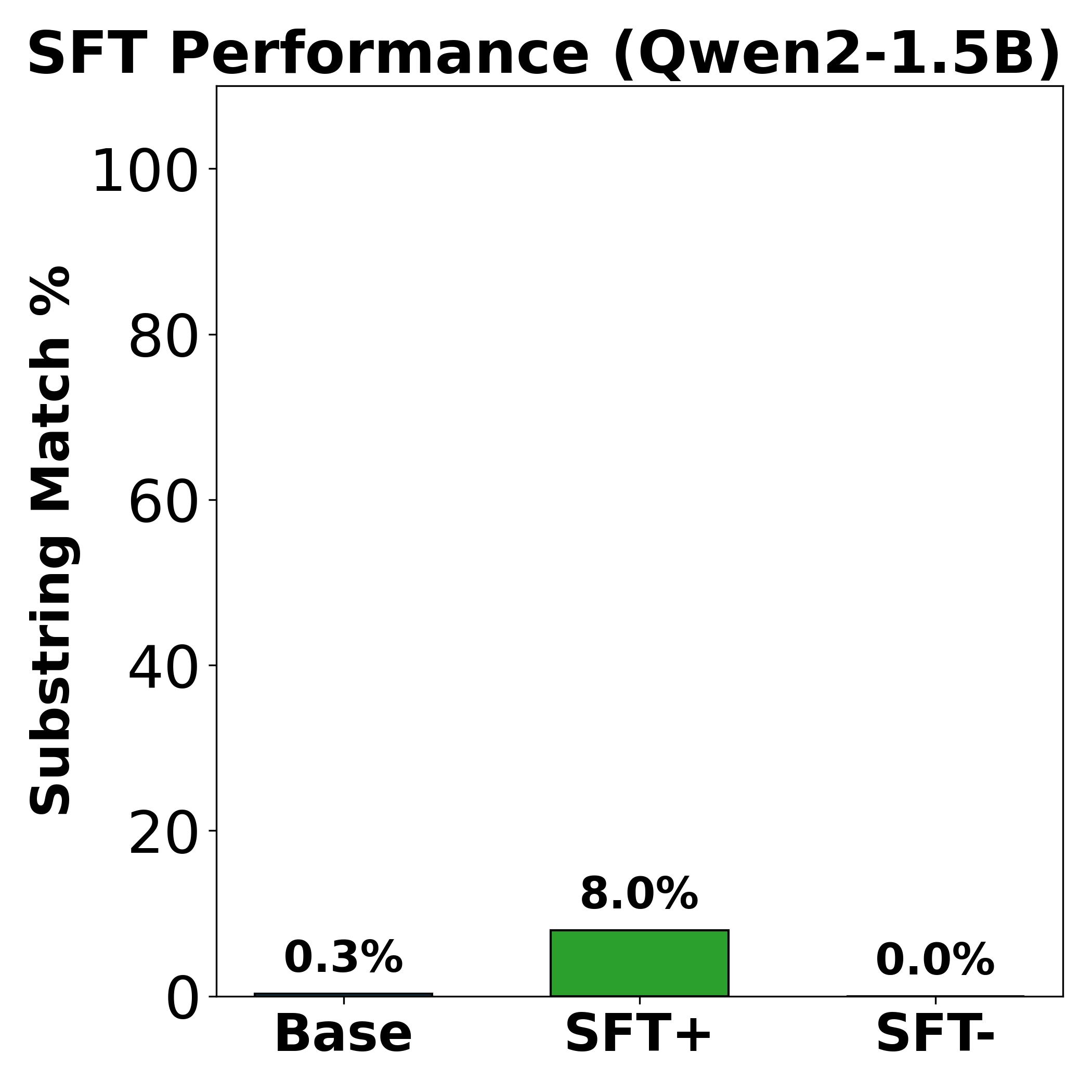}
    \hfill
    \includegraphics[width=0.6\linewidth]{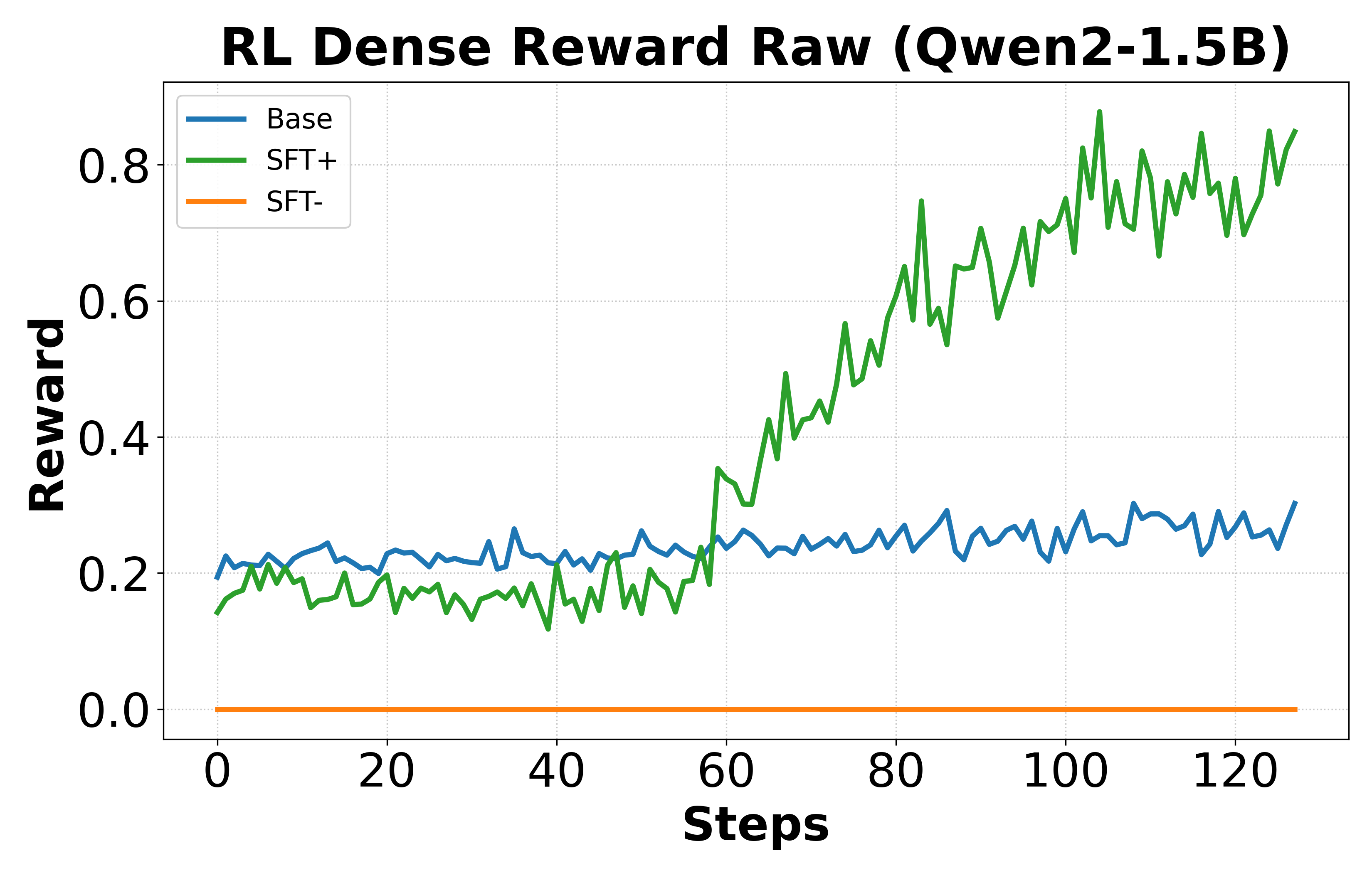}
    \hfill
    \includegraphics[width=0.38\linewidth]{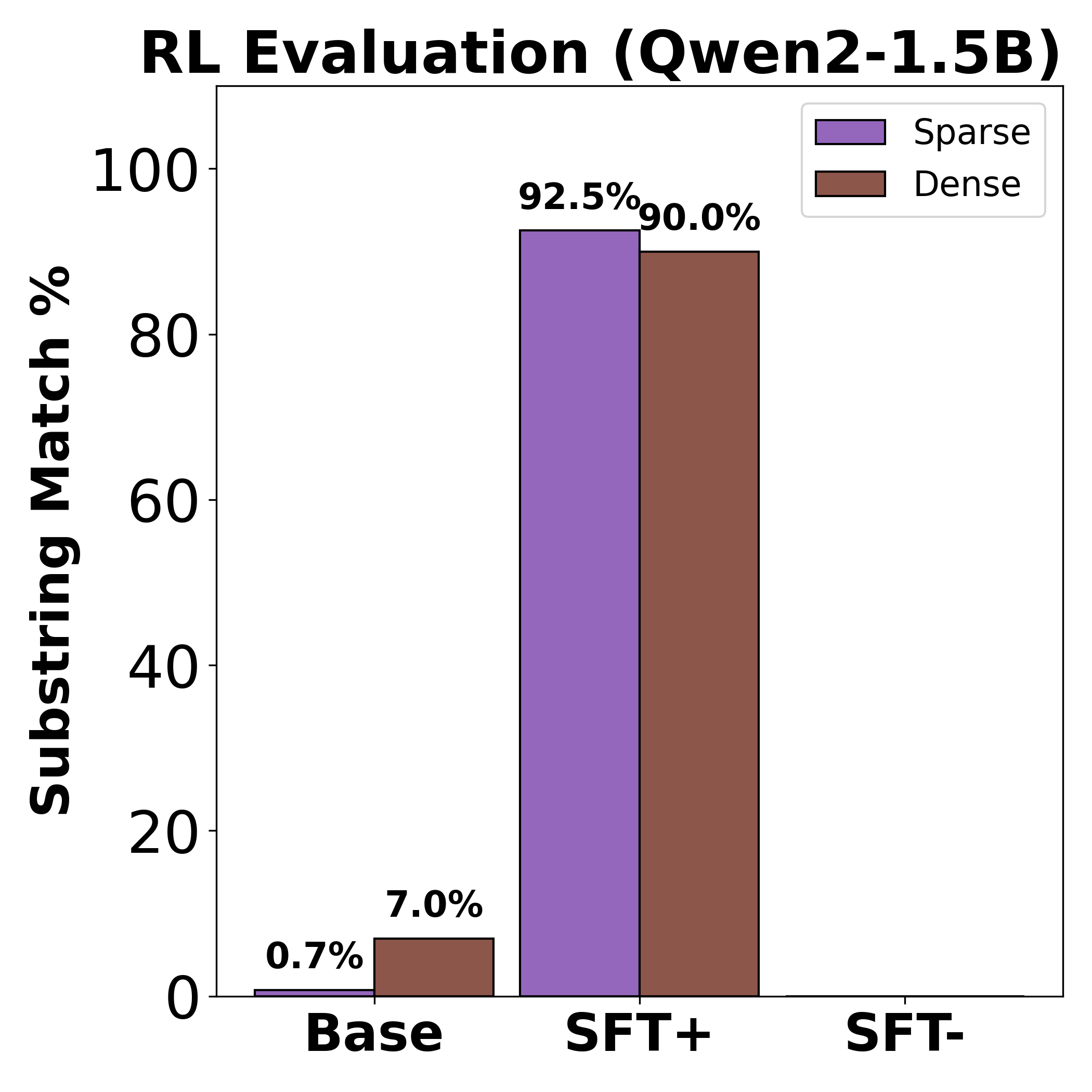}
    \hfill
    \includegraphics[width=0.6\linewidth]{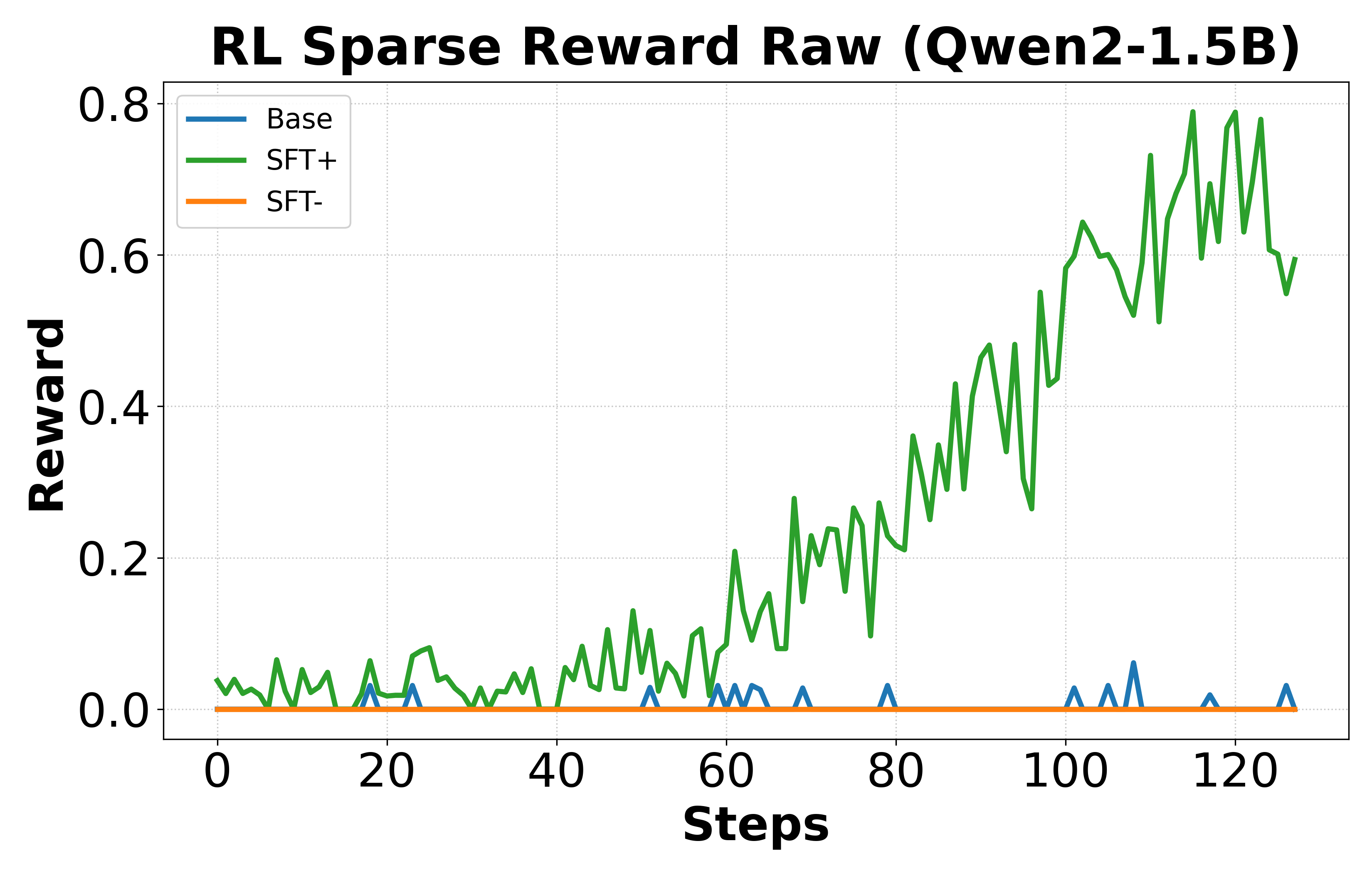}
    \hfill
    \caption{\textbf{RL post-training dynamics and evaluation metrics for Qwen2-1.5B.} \textbf{(Top Left)} Initial substring match performance prior to RL shows the SFT+ model possesses only marginal probability mass for the target sequence (8.0\%), while Base (0.3\%) and SFT- (0.0\%) exhibit near-zero coverage. \textbf{(Bottom Left)} Final RL evaluation demonstrates that only the SFT+ model successfully acquires the target behavior (achieving $>90\%$ match rates under both reward regimes). The Base model gains a slight benefit from dense rewards (7.0\%) but largely fails, alongside SFT-. \textbf{(Right)} Raw reward curves confirm these dynamics: the SFT+ model successfully optimizes both signals, while the Base and SFT- variants flatline, failing to escape their initial exploration plateaus.}
    \label{fig:qwen2-1.5b_Apdx}
\end{figure}

\begin{figure}[ht]
    \centering
    \noindent
    \includegraphics[width=0.38\linewidth]{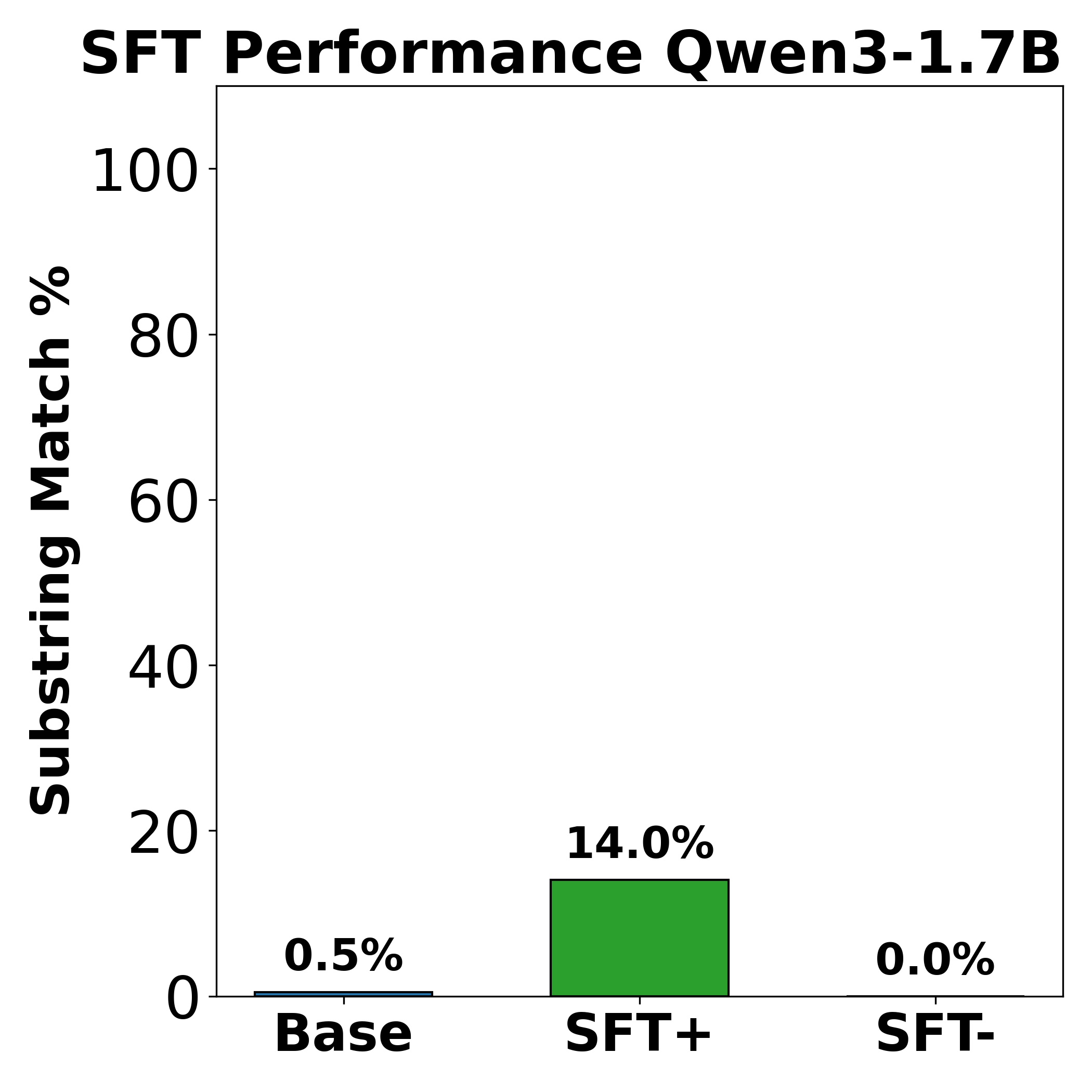}
    \hfill
    \includegraphics[width=0.6\linewidth]{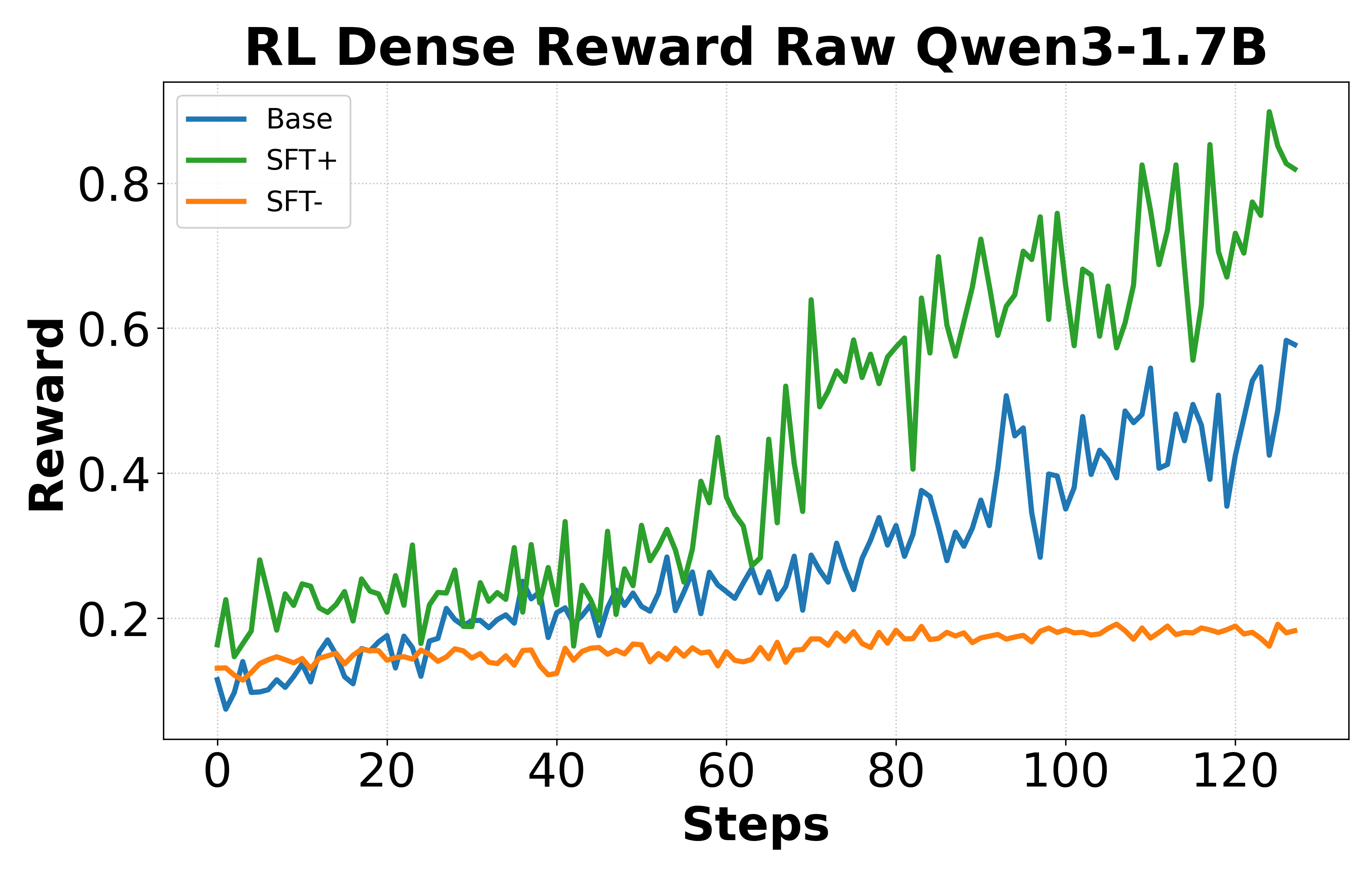}
    \hfill
    \includegraphics[width=0.38\linewidth]{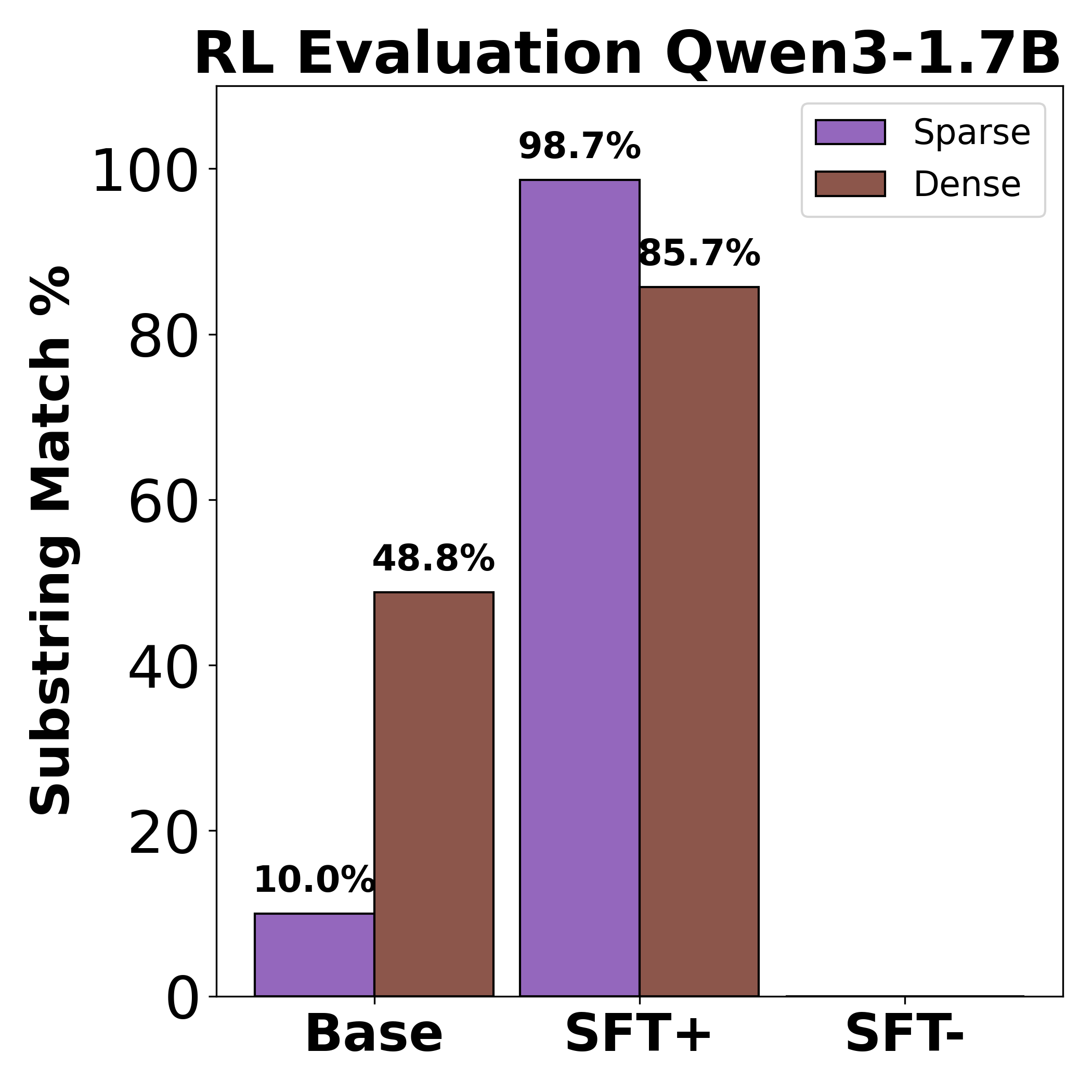}
    \hfill
    \includegraphics[width=0.6\linewidth]{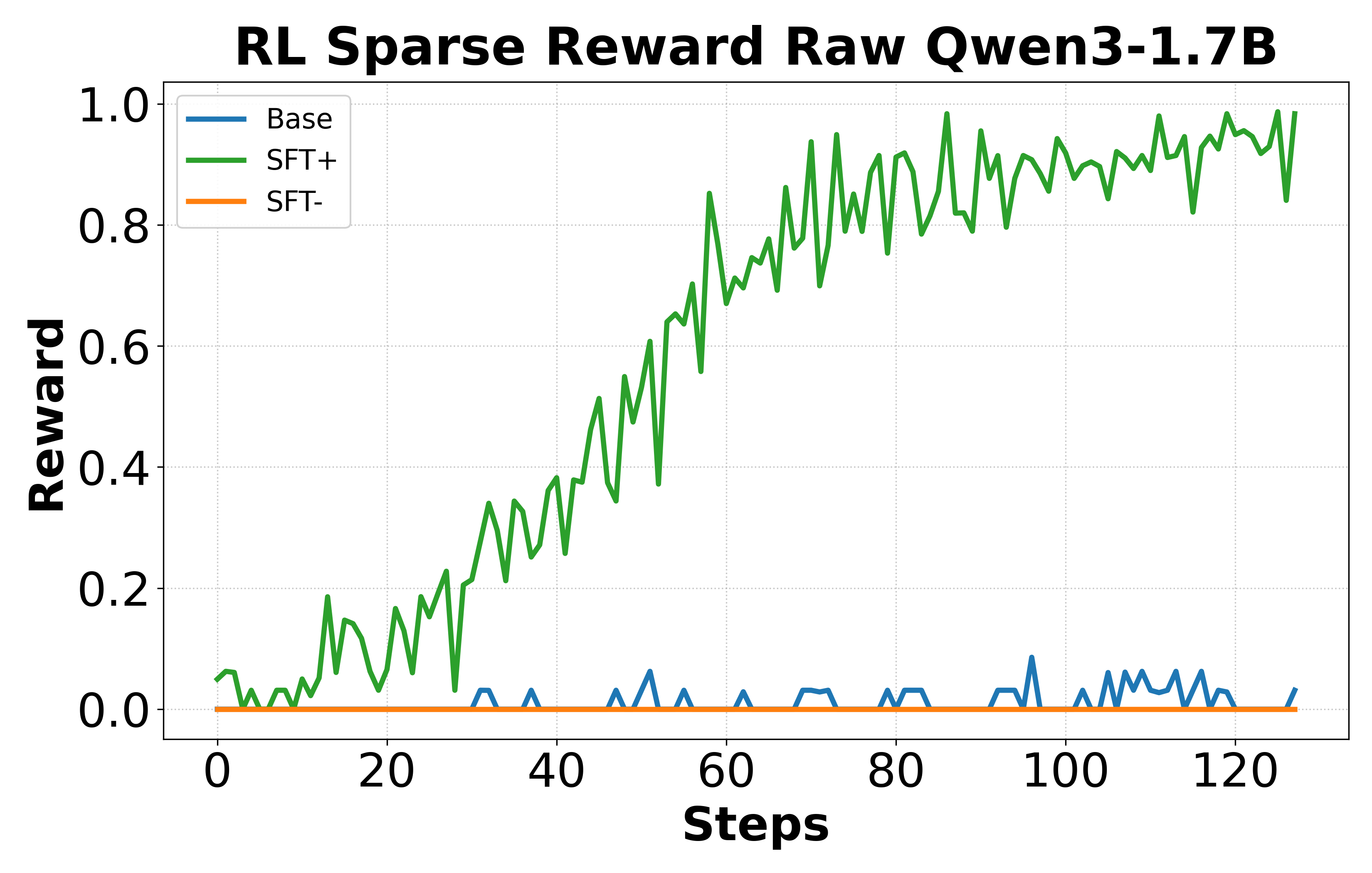}
    \hfill
    \caption{\textbf{RL post-training dynamics and evaluation metrics for Qwen3-1.7B.} \textbf{(Top Left)} Initial coverage is limited, with SFT+ at 14.0\% and Base/SFT- near zero. \textbf{(Bottom Left)} Final evaluation demonstrates the specific power of dense rewards. While the Base model only achieves a 10.0\% match rate under sparse rewards, the granular gradient of the dense reward allows it to reach 48.8\%. The SFT+ model succeeds broadly across both signals. \textbf{(Right)} The reward curves explicitly show this dynamic: under dense rewards, the Base model successfully begins climbing (reaching $\sim$0.5), whereas under sparse rewards, it remains completely flat at 0. SFT- fails entirely across all regimes.}
    \label{fig:qwen3-1.7b_Apdx}
\end{figure}

\begin{figure}[ht]
    \centering
    \noindent
    \includegraphics[width=0.38\linewidth]{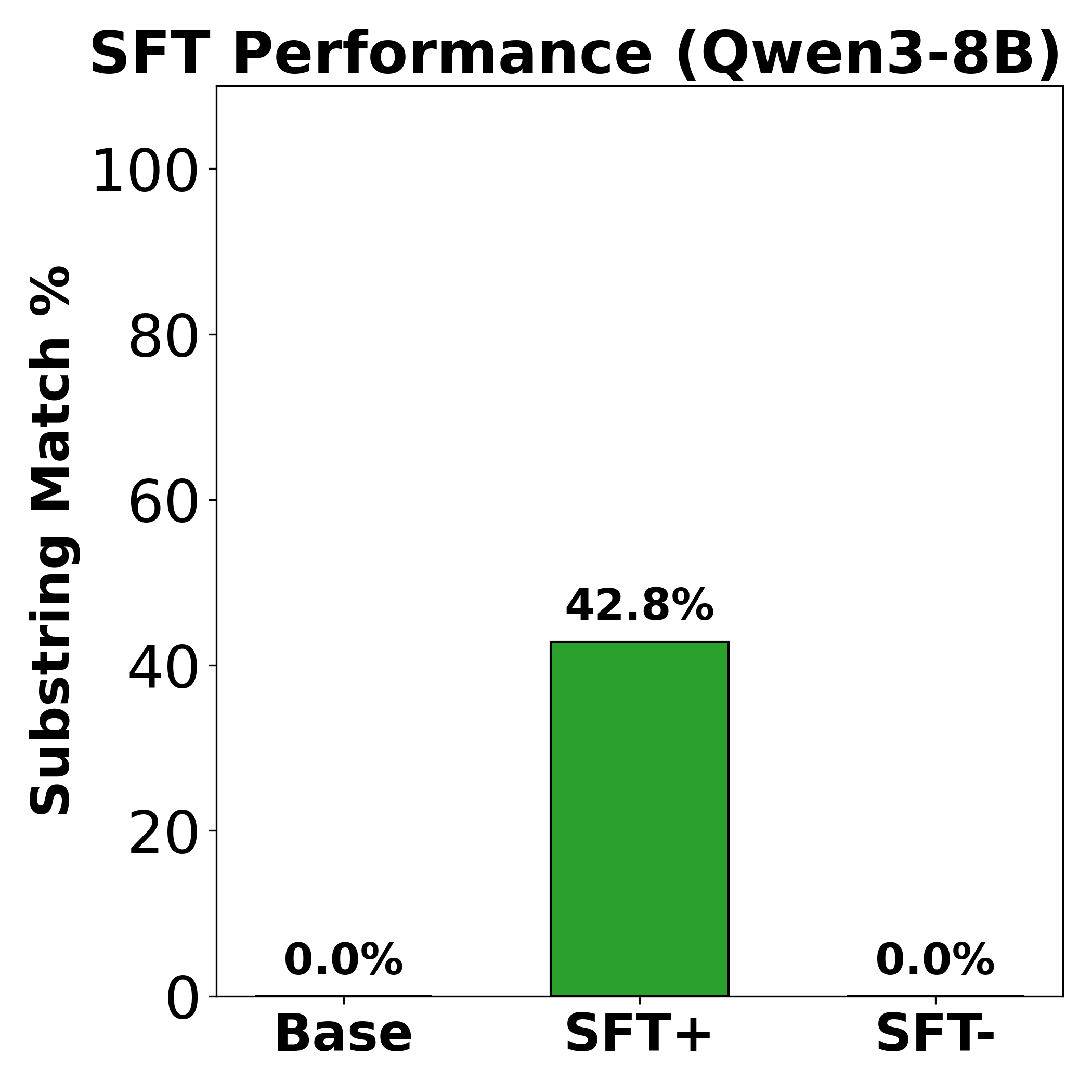}
    \hfill
    \includegraphics[width=0.6\linewidth]{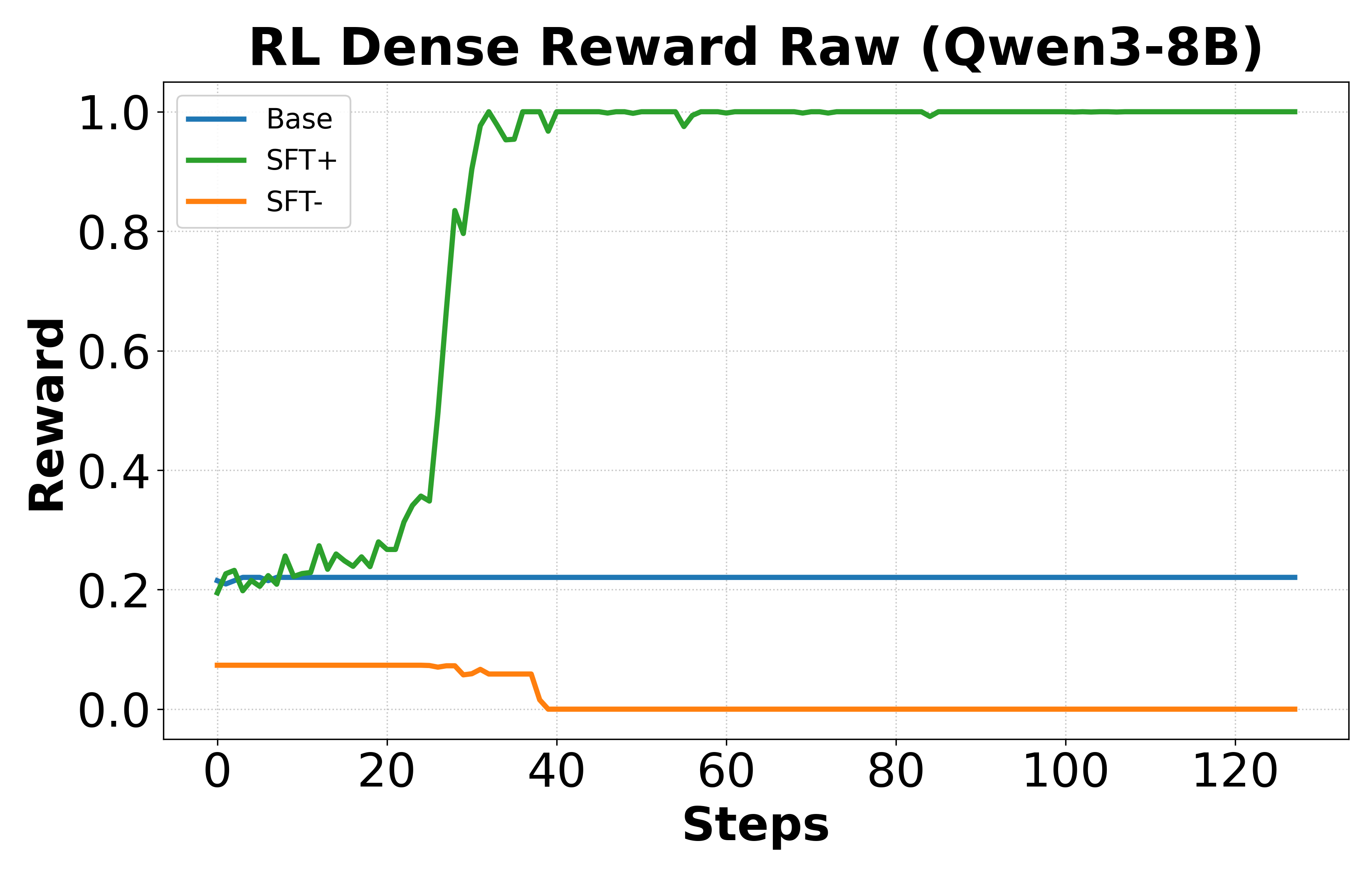}
    \hfill
    \includegraphics[width=0.38\linewidth]{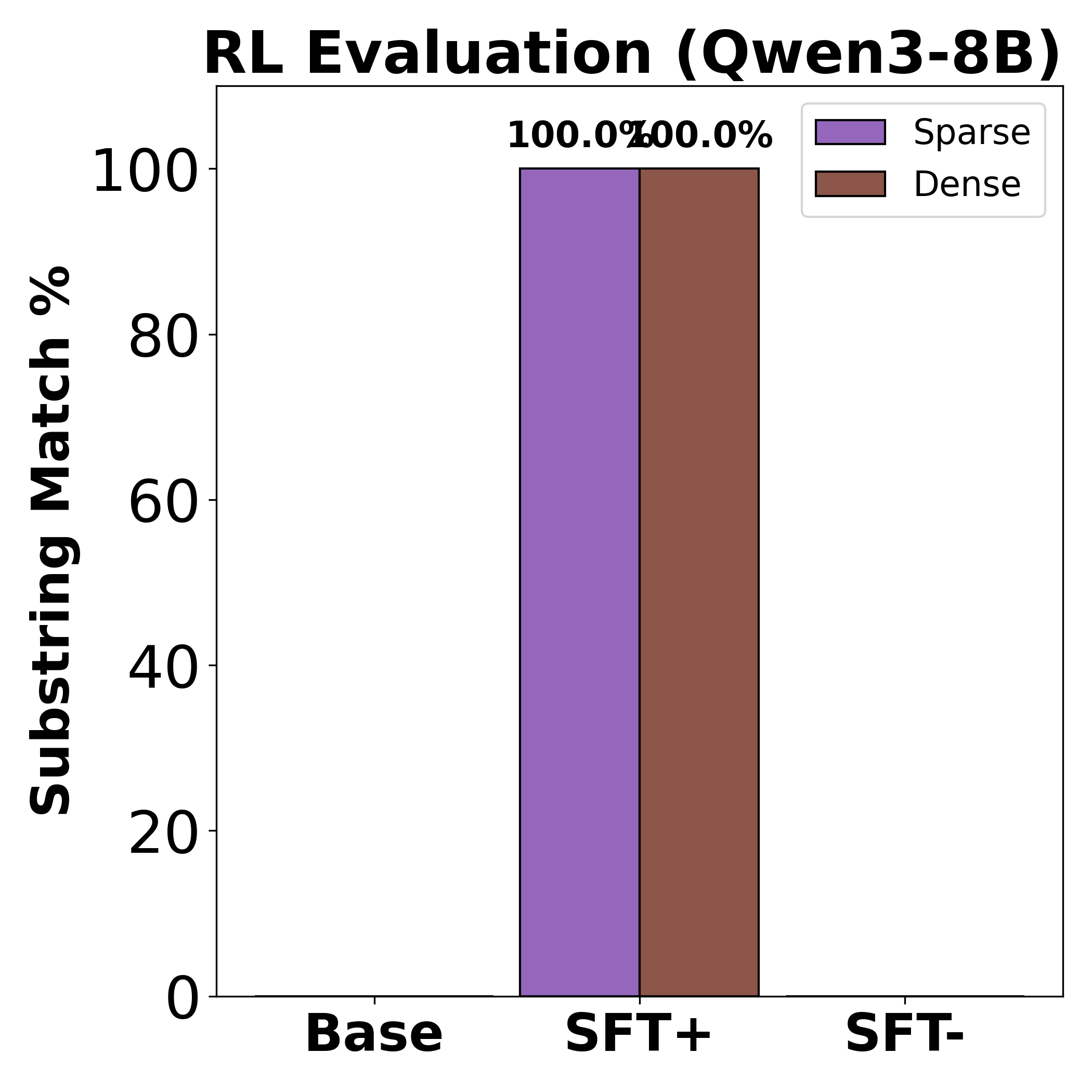}
    \hfill
    \includegraphics[width=0.6\linewidth]{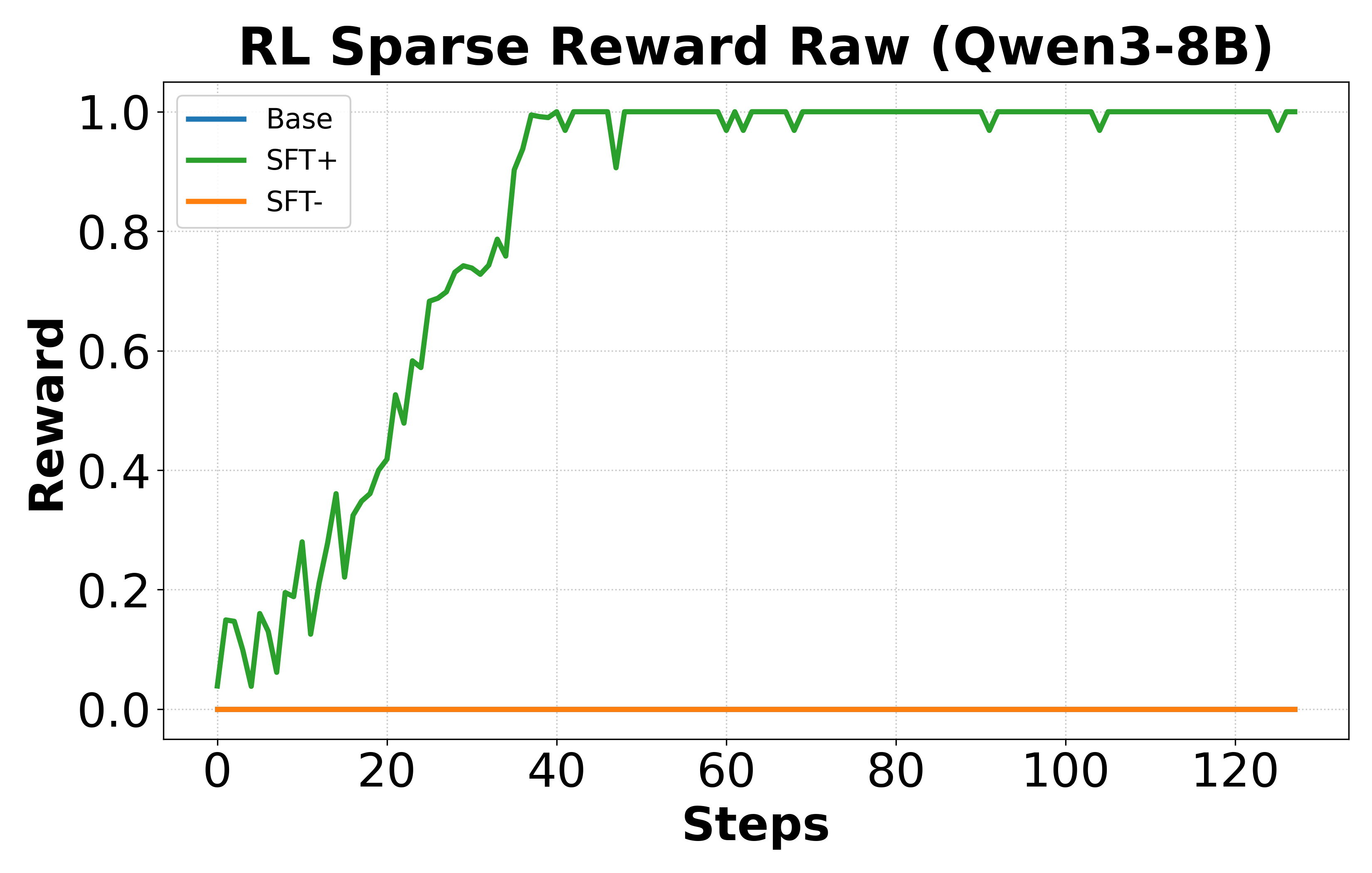}
    \hfill
    \caption{\textbf{RL post-training dynamics and evaluation metrics for Qwen3-8B.} \textbf{(Top Left)} The SFT+ model starts with a highly accessible prior (42.8\%), while the Base and SFT- models start with exactly 0.0\% probability mass. \textbf{(Bottom Left)} Final evaluation shows a stark binary outcome: the SFT+ model perfectly converges to a 100.0\% match rate under both sparse and dense rewards, while the Base and SFT- models achieve absolute 0\%. \textbf{(Right)} The raw reward curves demonstrate rapid, smooth convergence for SFT+ under both regimes. In contrast, the Base and SFT- models remain completely flat, unable to generate the target sequence a single time during exploration, confirming the strict bottleneck imposed by a 0\% base prior.}
    \label{fig:qwen3-8b_Apdx}
\end{figure}

\end{document}